\pdfoutput=1
\documentclass{article}

\usepackage[utf8]{inputenc} %
\usepackage[T1]{fontenc}    %

\usepackage[bitstream-charter]{mathdesign}
\usepackage{amsmath}
\usepackage[scaled=0.92]{PTSans}

\usepackage[
  paper  = letterpaper,
  left   = 1.65in,
  right  = 1.65in,
  top    = 1.0in,
  bottom = 1.0in,
  ]{geometry}

\usepackage[usenames,dvipsnames,table]{xcolor}
\definecolor{shadecolor}{gray}{0.9}

\usepackage[final,expansion=alltext]{microtype}
\usepackage[english]{babel}
\usepackage[parfill]{parskip}
\usepackage{afterpage}
\usepackage{framed}

{\endMakeFramed}

\usepackage{lineno}

\usepackage{ragged2e}

\newcommand{\parnum}{\bfseries\P\arabic{parcount}}
\newcounter{parcount}
\newcommand\p{%
    \stepcounter{parcount}%
    \leavevmode\marginpar[\hfill\parnum]{\parnum}%
}
\usepackage{graphicx}
\usepackage{wrapfig}
\usepackage[labelfont=bf,font=footnotesize,width=.9\textwidth]{caption}
\usepackage[format=hang]{subcaption}

\usepackage{booktabs,multirow,multicol}       %

\usepackage[algoruled,algo2e]{algorithm2e}
\usepackage{listings}
\usepackage{fancyvrb}
\fvset{fontsize=\normalsize}

\usepackage[colorlinks,linktoc=all]{hyperref}
\usepackage[all]{hypcap}
\hypersetup{citecolor=MidnightBlue}
\hypersetup{linkcolor=MidnightBlue}
\hypersetup{urlcolor=MidnightBlue}

\usepackage[nameinlink]{cleveref}

\usepackage[acronym,nowarn]{glossaries} %
\lstdefinestyle{mystyle}{
    commentstyle=\color{OliveGreen},
    keywordstyle=\color{BurntOrange},
    numberstyle=\tiny\color{black!60},
    stringstyle=\color{MidnightBlue},
    basicstyle=\ttfamily,
    breakatwhitespace=false,
    breaklines=true,
    captionpos=b,
    keepspaces=true,
    numbers=left,
    numbersep=5pt,
    showspaces=false,
    showstringspaces=false,
    showtabs=false,
    tabsize=2
}
\usepackage{amsthm}

\usepackage{centernot}
\usepackage{nicefrac}       %
\usepackage{mathtools}
\usepackage{amsbsy}
\usepackage{amstext}
\usepackage{thmtools}
\usepackage{thm-restate}

\begingroup
    \makeatletter
    \@for\theoremstyle:=definition,remark,plain\do{%
        \expandafter\g@addto@macro\csname th@\theoremstyle\endcsname{%
            \addtolength\thm@preskip\parskip
            }%
        }
\endgroup

\crefname{lemma}{lemma}{lemmas}
\Crefname{lemma}{Lemma}{Lemmas}
\crefname{thm}{theorem}{theorems}
\Crefname{thm}{Theorem}{Theorems}
\crefname{prop}{proposition}{propositions}
\Crefname{prop}{Proposition}{Propositions}
\crefname{assumption}{assumption}{assumptions}
\crefname{assumption}{Assumption}{Assumptions}

\renewcommand{\mid}{~\vert~}

\newcommand{\V}{\mathbb{V}}

\newcommand{\g}{\, | \,}
\newcommand{\s}{\, ; \,}

\newcommand{\E}[2]{\mathbb{E}_{#1}\left[#2\right]}

\usepackage{booktabs,arydshln}
\makeatletter
\def\adl@drawiv#1#2#3{%
        \hskip.5\tabcolsep
        \xleaders#3{#2.5\@tempdimb #1{1}#2.5\@tempdimb}%
                #2\z@ plus1fil minus1fil\relax
        \hskip.5\tabcolsep}
\newcommand{\cdashlinelr}[1]{%
  \noalign{\vskip\aboverulesep
           \global\let\@dashdrawstore\adl@draw
           \global\let\adl@draw\adl@drawiv}
  \cdashline{#1}
  \noalign{\global\let\adl@draw\@dashdrawstore
           \vskip\belowrulesep}}
\makeatother

\renewcommand{\epsilon}{\varepsilon}

\declaretheorem[style=plain,name=Theorem]{theorem}

\declaretheorem[style=definition,name=Assumption]{assumption}
\declaretheorem[style=definition,sibling=theorem,name=Example]{example}

\newenvironment{example*}
 {\pushQED{\qed}\example}
 {\popQED\endexample}
\numberwithin{equation}{section}

\newcommand{\defeq}{\coloneqq}

\DeclareMathOperator*{\logit}{logit}

\newcommand{\EE}{\mathbb{E}}

\newcommand{\given}{\mid}

\providecommand\given{} %
\newcommand\SetSymbol[1][]{
  \nonscript\,#1:\nonscript\,\mathopen{}\allowbreak}
\DeclarePairedDelimiterX\Set[1]{\lbrace}{\rbrace}%
{ \renewcommand\given{\SetSymbol[]} #1 }

\usepackage{subcaption}
\usepackage{bm}

\usepackage{csquotes}
\usepackage[%
minnames=1,maxnames=99,maxcitenames=2,
style=alphabetic,
doi=false,
url=false,
giveninits=true,
hyperref,
natbib,
backend=bibtex,
sorting=nyt,
backref=true
]{biblatex}%
\renewbibmacro{in:}{%
  \ifentrytype{article}{}{\printtext{\bibstring{in}\intitlepunct}}}

\renewbibmacro*{journal}{%
  \iffieldundef{journaltitle}
    {}
    {\printtext[journaltitle]{%
       \printfield[noformat]{journaltitle}%
       \setunit{\subtitlepunct}%
       \printfield[noformat]{journalsubtitle}}}}

\DeclareFieldFormat{sentencecase}{\MakeSentenceCase*{#1}}

\renewbibmacro*{title}{%
  \ifthenelse{\iffieldundef{title}\AND\iffieldundef{subtitle}}
    {}
    {\ifthenelse{\ifentrytype{article}\OR\ifentrytype{inbook}%
      \OR\ifentrytype{incollection}\OR\ifentrytype{inproceedings}%
      \OR\ifentrytype{inreference}\OR\ifentrytype{misc}}
      {\printtext[title]{%
        \printfield[sentencecase]{title}%
        \setunit{\subtitlepunct}%
        \printfield[sentencecase]{subtitle}}}%
      {\printtext[title]{%
        \printfield[titlecase]{title}%
        \setunit{\subtitlepunct}%
        \printfield[titlecase]{subtitle}}}%
     \newunit}%
  \printfield{titleaddon}}

\AtEveryBibitem{%
\ifentrytype{article}{
    \clearfield{urldate}%
    \clearfield{eprint}
    \clearfield{eid}
}{}
\ifentrytype{book}{
    \clearfield{url}%
    \clearfield{urldate}%
    \clearfield{eprint}
}{}
\ifentrytype{collection}{
    \clearfield{url}%
    \clearfield{urldate}%
    \clearfield{eprint}
}{}
\ifentrytype{incollection}{
    \clearfield{url}%
    \clearfield{urldate}%
    \clearfield{eprint}
}{}
}

\AtEveryBibitem{
    \clearfield{pages}
    \clearfield{review}%
    \clearfield{series}%
    \clearfield{volume}
    \clearfield{month}
    \clearfield{isbn}
    \clearfield{issn}
    \clearlist{location}
    \clearfield{series}
    \clearlist{publisher}
    \clearname{editor}
}{}

\crefformat{equation}{(#2#1#3)}
\crefformat{figure}{Figure~#2#1#3}
\crefname{example}{Example}{Examples}
\crefname{lemma}{Lemma}{Lemmas}
\crefname{cor}{Corollary}{Corollaries}
\crefname{theorem}{Theorem}{Theorems}
\crefname{assumption}{Assumption}{Assumptions}

\usepackage{enumitem} %
\usepackage[separate-uncertainty=true,multi-part-units=single]{siunitx} %

\declaretheoremstyle[
spacebelow=\parsep,
    spaceabove=\parsep,
  mdframed={
    backgroundcolor=gray!10!white,     %
    hidealllines=true, 
    innertopmargin=8pt, 
    innerbottommargin=4pt, 
    skipabove=8pt,
    skipbelow=10pt,
    nobreak=true
}
]{grayboxed}
\declaretheorem[style=grayboxed,name=Theorem]{gtheorem}
\crefname{gtheorem}{Theorem}{Theorems}

\usepackage{thm-restate}

\usepackage{xcolor}

\definecolor{WowColor}{rgb}{.75,0,.75}
\definecolor{SubtleColor}{rgb}{0,0,.50}

\newcounter{margincounter}

\usepackage[affil-it]{authblk}

\title{Transforming and Combining Rewards \\for Aligning Large Language Models}
\date{}
\author[1]{Zihao Wang\thanks{Work done during internship at Google DeepMind}}
\author[2]{Chirag Nagpal}
\author[3]{Jonathan Berant}
\author[3]{Jacob Eisenstein}
\author[3]{Alex D'Amour}
\author[4,3]{Sanmi Koyejo}
\author[1,3]{Victor Veitch}
\affil[1]{University of Chicago}
\affil[2]{Google Research}
\affil[3]{Google DeepMind}
\affil[4]{Stanford University}
\begin{document}
\maketitle
\renewcommand\thefootnote{} \footnotetext{Published in ICML 2024.}

\begin{abstract}
A common approach for aligning language models to human preferences is to first learn a reward model from preference data, and then use this reward model to update the language model.
We study two closely related problems that arise in this approach.
First, any monotone transformation of the reward model preserves preference ranking; is there a choice that is ``better'' than others?
Second, we often wish to align language models to multiple properties: how should we combine multiple reward models?
Using a probabilistic interpretation of the alignment procedure, we identify a natural choice for transformation for (the common case of) rewards learned from Bradley-Terry preference models. 
The derived transformation is straightforward: we apply a log-sigmoid function to the centered rewards, a method we term ``LSC-transformation'' (log-sigmoid-centered transformation). 
This transformation has two important properties. 
First, it emphasizes improving poorly-performing outputs, rather than outputs that already score well. This mitigates both underfitting (where some prompts are not improved) and reward hacking (where the model learns to exploit misspecification of the reward model).
Second, it enables principled aggregation of rewards by linking summation to logical conjunction: the sum of transformed rewards corresponds to the probability that the output is ``good'' in all measured properties, in a sense we make precise. 
Experiments aligning language models to be both helpful and harmless using RLHF show substantial improvements over the baseline (non-transformed) approach.
\end{abstract}

\section{Introduction}\label{sec:intro}
In this paper, we are interested in how to align large language models in order to bias their outputs towards having desired properties---e.g., to be helpful, harmless, factual, or creative \cite{ziegler2019fine,stiennon2020learning,ouyang2022training}.
We study the two-stage approach where we first learn a reward model from human preferences, and then align the language model so that its outputs have high values under the reward model.
In this context, we're interested in two fundamental problems:
\begin{enumerate}
  \item The alignment step maximizes the expected learned reward model.
  However, any monotone transformation of the reward preserves the interpretation of the alignment procedure as biasing the language model towards human-preferred outputs. Can we improve the alignment step by transforming the learned reward? 
  \item We often wish to align language models to multiple properties---e.g., outputs should be helpful, and harmless, and factual. If we have reward models for each property, how should we combine them? 
\end{enumerate}

A main challenge in answering these questions is that the goal of alignment is not precisely defined. As a result, there is no obvious principle to guide the choice of transformation or aggregation method.
The conceptual idea in this paper is to interpret alignment probabilistically.
From this perspective, the goal of aligning a model to a particular property is to produce samples from the posterior distribution conditional on the outputs being ``good'' on that property.
Similarly, the goal of aligning to multiple properties is to produce samples conditional on the outputs being ``good'' on all properties.

To make use of this idea, we need to define what it means for an output to be ``good''.
In the context of rewards learned from preference data, we take an output $y$ be ``good'' if it has reward $r(x, y)$ greater than some prompt-specific reference value $r^\text{ref}(x)$. 
The first main result of the paper is that in the (typical) case where the reward model is learned from preference data using a Bradley-Terry model and the language model is aligned by maximizing expected reward subject to a KL constraint, the natural choice of transformation is:\looseness=-1
\begin{align}
  u(x, y) = \log \sigma(r(x, y) - r^\text{ref}(x)),
\end{align}
where $\sigma(\cdot)$ is the sigmoid function.
Here, $r$ is the learned Bradley-Terry reward model, and $u$ is the transformed reward we use in the alignment step.
We call this transformation the ``LSC-transformation'' (log-sigmoid-centered transformation), and alignment using such transformation ``LSC-alignment''.

This transformation is motivated by a probabilistic interpretation. 
It additionally turns out to have important practical benefits relative to the baseline approach of using the raw reward model. 
First, the transformed reward shrinks the marginal utility of very high reward values. This has the effect in alignment of both encouraging the model to improve poorly performing prompts, and of discouraging the model from ``reward hacking'' by optimizing the reward model outside the range of its validity.
Second, the transformed reward offers a natural way to combine multiple reward models. 
Namely: the sum of the transformed rewards corresponds to the logical AND of the outputs being ``good'' on each property.
So, after transforming the rewards, we can simply sum them to aggregate multiple reward models.\looseness=-1

\begin{figure}[t]
    \centering
    \includegraphics[width=0.7\textwidth]{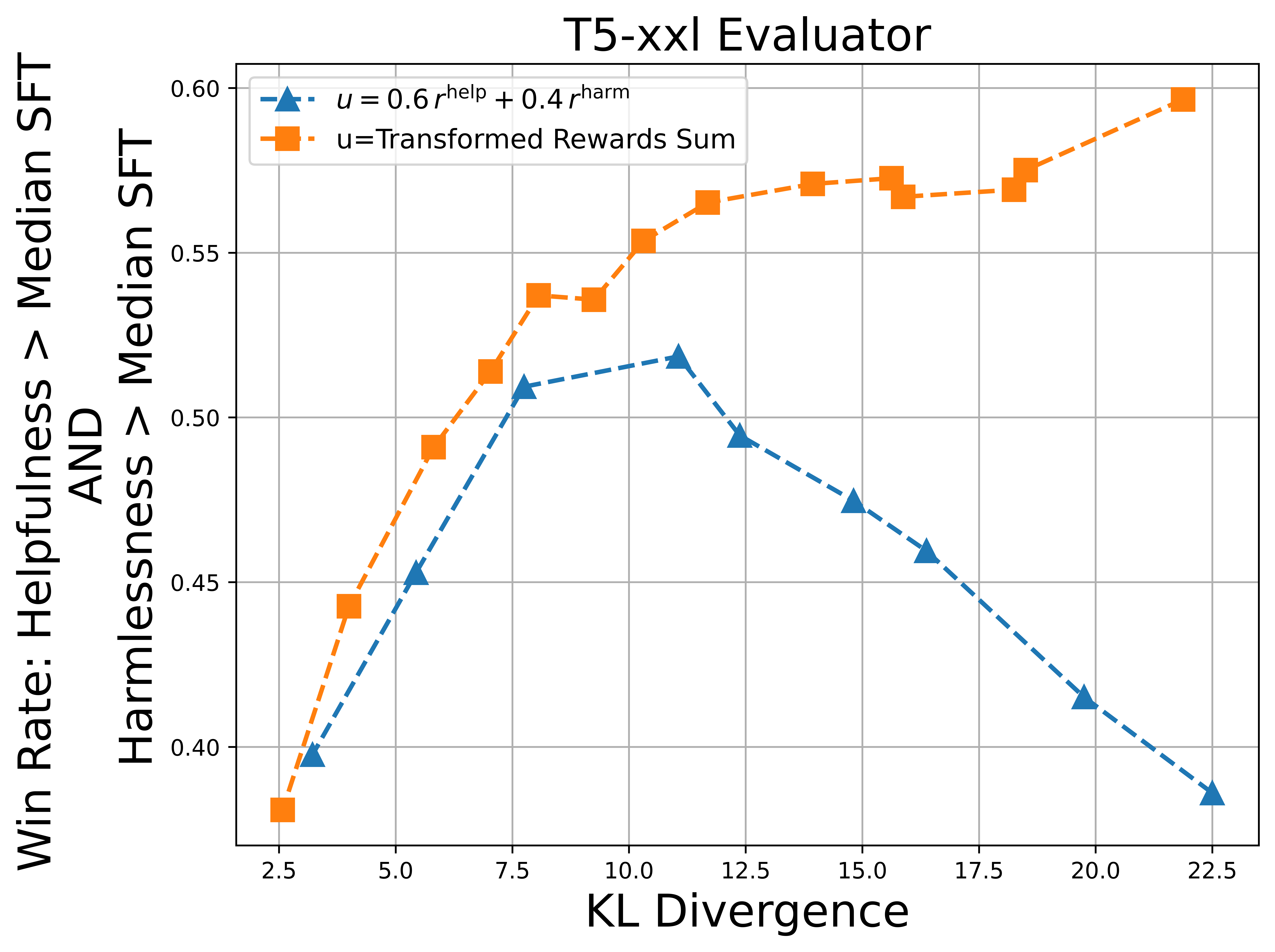}
    \caption{Transforming the Bradley-Terry reward both mitigates overfitting and makes addition behave as logical AND. This leads to significant improvements in aligned model quality relative to standard practice.
    Each point on the plot is a LLM aligned with a different KL penalty weight.
    The $y$-axis shows improvement over the base supervise finetuned (SFT) LLM in both helpfulness AND harmlessness, as judged by an external evaluator model (not used for RLHF). The baseline aggregates suboptimally (usually losing on either helpfulness or harmlessness) and suffers reward hacking (performance decays in the high KL regime). Details in \cref{sec:experiments}.%
    }
    \label{fig:wr_agg_selected}
    \end{figure}
    
In combination, these benefits can lead to substantial improvements in alignment performance. \Cref{fig:wr_agg_selected} compares aligning a language model to be both helpful and harmless using summation of transformed and untransformed rewards. 
Varying the strength of KL regularization used in the alignment step, we observe that the transformed reward leads to substantial improvements at all KL levels.  

\section{Preliminaries}
We first review the standard Reinforcement Learning from Human Feedback (RLHF) two-step procedure for aligning language models to human preferences. 

\paragraph{Reward model training from pairwise data}
Reward models are trained to emulate human feedback. A frequently used type of feedback is pairwise preference data, consisting of a prompt $x$ and two generated responses $y^+, y^-$, where $y^+$ is preferred by the human annotator.  Our discussion mainly focuses on this case.

Commonly, rewards are learned using the Bradley-Terry model \cite{bradley1952rank},
\begin{align}
p(y^- \prec y^+ \mid x) = \sigma(r(x, y^+) - r(x, y^-)).
\label{eq:bt}
\end{align}
The function $r$ is parameterized by a neural network (typically, another LLM) and fit using the standard maximum log likelihood objective.

\paragraph{Alignment to reward model}
The next step is updating the base LLM to bias it towards high-reward responses.

Usually, aligning the model to the reward function is proceeded by a ``Supervised Finetuning'' (SFT) step where the base model is fine-tuned using the language modeling objective on the winning examples from the human preference data.
We denote this model as $\pi_0$. Our interest is how to use the reward model to further align $\pi_0$.

The aim of the alignment step is to update $\pi_0$ to a new model $\pi^*$ that has high expected reward, while still being close to $\pi_0$ (to preserve information from the previous training phases). %
Standard practice is to learn $\pi^*$ by maximizing the expected reward of samples, regularized by a penalty on the KL-divergence between $\pi^*$ and $\pi_0$.

The main idea in this paper is to instead use a utility measure $u(x,y)$ that is a monotone transformation of $r(x,y)$. 
We leave the alignment procedure otherwise unchanged.
Then, mathematically, $\pi^*$ is the maximizer of:
\begin{align}
    \EE_{x} \{ \EE_{y \sim \pi(\cdot| x)} [u(x, y)] - \gamma \text{KL}(\pi(\cdot | x) \| \pi_{0}(\cdot | x)) \} \label{eq:rlhf_obj}
\end{align}
Here, $\gamma$ is a hyper-parameter that controls the trade-off between maximizing rewards and aligning with $\pi_0$. %

\section{Reward Transformation}
\label{sec:methodology-transform}
\begin{figure}[t]
  \centering
  \begin{subfigure}[t]{0.4\columnwidth}
    \includegraphics[width=\linewidth]{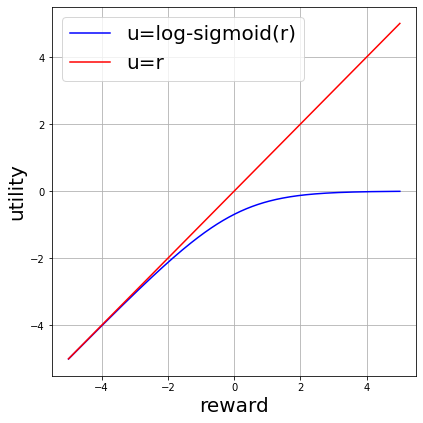}
    \caption{Transformation Shape}
    \label{fig:shape}
  \end{subfigure}%
  \hfill
  \begin{subfigure}[t]{0.5\columnwidth}
    \includegraphics[width=\linewidth]{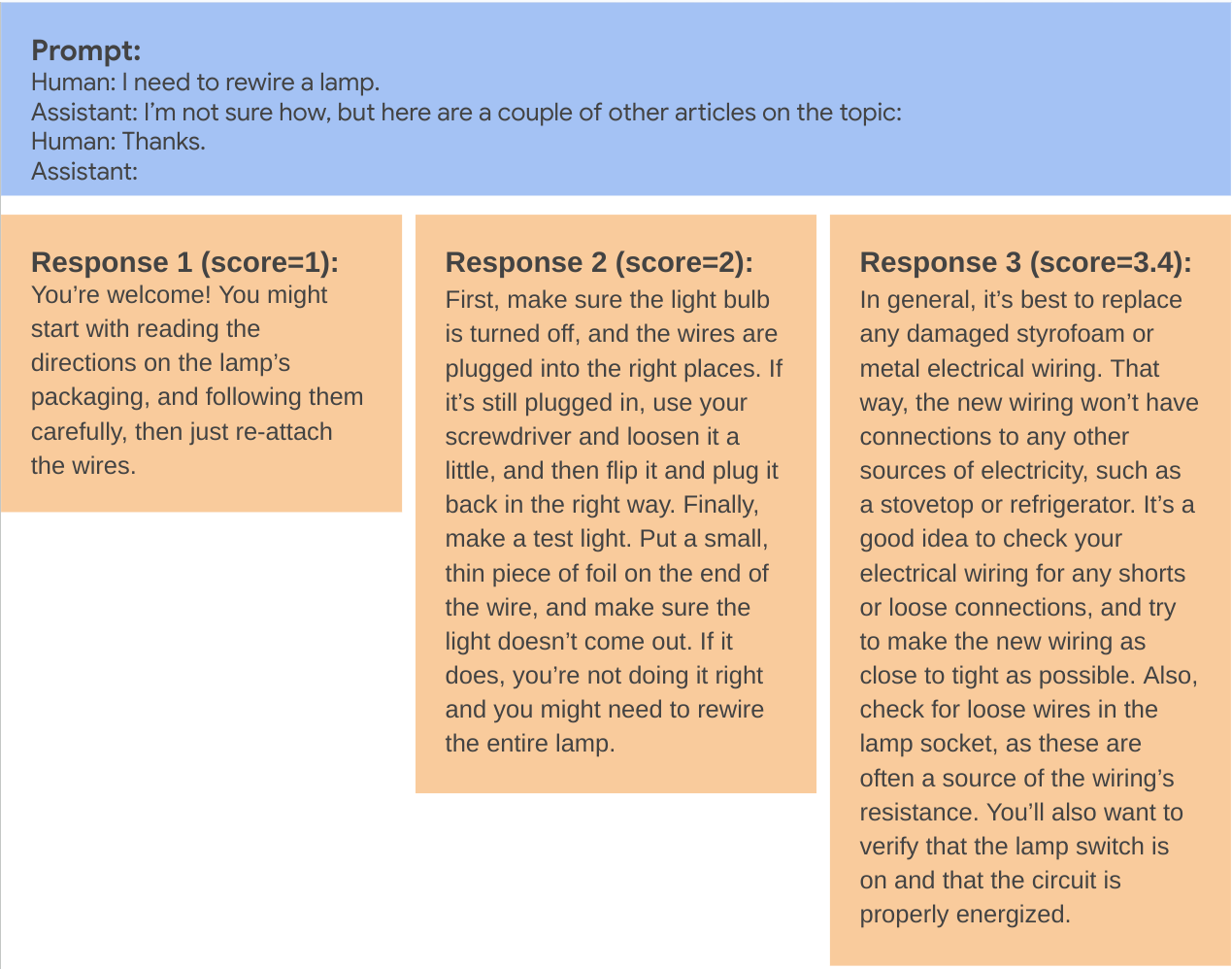}
    \caption{Helpfulness Examples}
    \label{fig:example}
  \end{subfigure}
  \caption{Bradley-Terry rewards do not capture diminishing utility, and log-sigmoid transforming can fix this.
  In the example responses, moving from response 1 to response 2 substantially increases utility, but from response 2 to response 3 only marginally increases. However, the BT rewards treat each improvement the same. A log-sigmoid transformation reflects diminishing returns.  
  }
  \label{fig:shape_and_examples}
\end{figure}
We now turn to deriving the reward transformation. 

\paragraph{Formalize Alignment Goal}
The first step is to formalize the goal of alignment. This is necessary to identify a ``correct'' transformation.
Intuitively we want to modify the initial policy $\pi_0(y|x)$ so that the generated samples are considered ``good'' in some property by humans. 
To make progress, we introduce a binary semantic random variable $G$ indicating whether response $y$ is ``good'' for prompt $x$. 
Then we define the alignment goal as producing a model that samples from the distribution of responses conditional on the response being good; i.e., $\pi^\mathrm{target}(\cdot \given x) = p(\cdot \given x, G=1)$. 

In fact, we slightly generalize this to allow finer grained control of the reward vs KL tradeoff.
By Bayes' rule, we may rewrite 
$p(y|x, G=1) \propto \pi_0(y|x) p(G=1|x, y)$.
That is, we reweight the base LLM by a term that upweights responses that are likely to be deemed good.
It is natural to introduce a hyperparameter to control the strength of this upweighting. 
Anticipating the connection to \cref{eq:rlhf_obj}, we again use $\gamma$ for this hyperparameter.
Then, we define our alignment goal as producing an aligned model%
\begin{align}
    \pi^\mathrm{target}_\gamma(y|x) \propto \pi_0(y|x) p(G=1|x, y)^{1/\gamma}\label{eq:target_distn} 
\end{align}

\paragraph{Reward Transformation}
The next question is how to produce an aligned model that satisfies our goal.
This has two parts: we must use the reward model to define the binary goodness variable $G$, and we must determine how the utility function used for alignment relates to $G$. 

\paragraph{Target Utility Function}
We begin by connecting alignment utility and $G$.
The idea is to use the well-known result that the ideal optimizer of the KL-regularized RLHF objective \cref{eq:rlhf_obj} is an exponential tilting of the base policy \citep[e.g.,][]{korbak2022rl}:
\begin{align}
    \pi^{*}(y \given x) \propto \pi_0(y \given x) \exp{(u(x, y)/\gamma)}
    \label{eq:aligned_dist}
\end{align}

Comparing \cref{eq:target_distn} with \cref{eq:aligned_dist}, we see that in order to get the target policy through alignment, we must set the utility function to be the log-probability of goodness: 
\begin{align}
u(x, y) = \log p(G=1|x, y). 
\end{align}

\paragraph{Pointwise Reward Models} 
The next step is to relate the Bradley-Terry model to $p(G=1|x, y)$.
As a warmup, we consider the case of reward models trained on pointwise data; i.e., where each example is a prompt $x$, response $y$, and a binary label $G$ indicating whether the response is good.
In this case, we would train the reward function by minimizing a binary cross-entropy objective.
This is a proper scoring rule, so the learned reward would be:
$r(x, y) = \text{logit} \ p(G=1|x, y)$. 
Here, we take the reward to be on the logit scale so that $r(x,y) \in (-\infty,\infty)$, analogous to Bradley-Terry rewards.
In this case, the right utility is $u(x, y) = \log \sigma(r(x, y))$. 

\paragraph{Pairwise Reward Models} 
The pairwise case is more subtle.
It may be tempting to again apply the $\log(\sigma(\cdot))$ transformation to the Bradley-Terry rewards.
This is incorrect for two reasons.
First, only reward \emph{differences} are interpretable as logit probilities---the rewards for individual prompts are not.
Second, in general the reward model $r(x, y)$ is unidentifiable from the data. For any $r$, we can shift it by any arbitrary function of the prompt $x$ without changing the Bradley-Terry model. That is, $\tilde{r}(x, y) \leftarrow r(x, y) + C(x)$ has the same objective in \cref{eq:bt}.
However, any non-linear transformation of the reward will be sensitive to this unidentified $C(x)$. 

Happily, both problems are resolved by choosing a suitable definition of what it means for a response to be good. Here, we take a generated response $y$ to be ``good'' if it is  preferred over a chosen reference output $y^{\text{ref}}$. 
For example, we may say that $y$ is harmless as long as it is preferred by the harmlessness reward to a canned response such as ``I am unable to answer that question".

The LSC-transformation follows immediately:
\begin{gtheorem}\label{thm:transformation}
Suppose output $y$ is deemed good for prompt $x$ if it would be preferred to reference output $y^\text{ref}(x)$.
Then, if the Bradley-Terry model \cref{eq:bt} holds, and we align using KL-regularized utility maximization \cref{eq:rlhf_obj}, then using utility
\begin{equation}
    u(x, y) = \log \sigma(r(x, y) - r^\text{ref}(x))
\end{equation}
will satisfy the alignment goal \cref{eq:target_distn}.
Here $r^\text{ref}(x) := r(x, y^\text{ref}(x))$. 

\end{gtheorem}

That is, once we decide on a reference response, we get the right utility function by applying log-sigmoid transformation to the centered reward.%

\paragraph{Mechanistic Interpretation}
We derived the reward transformation from a probabilistic argument. It is also insightful to consider the mechanistic effect of the transformation.

One fundamental issue with the baseline approach, where $u(x, y) = r(x, y)$, is that the utility gain for improvements never diminishes. 
In the aligned model, taking $\gamma=1$, we have that the relative probabilities of two responses is exponential in the difference of utilities.
\begin{align}
\frac{\pi^*(y^1 \given x)}{\pi^*(y^0 \given x)} = \exp(u(y_1, x) - u(y_0, x)) \frac{\pi_0(y^1 \given x)}{\pi_0(y^0 \given x)}
\end{align}%
Now, consider the case where we have three candidate responses: $y^\text{ref}, y^0, y^1$ such that $p(y_\text{ref} \prec y^0) = 0.99$ and $p(y_\text{ref} \prec y^1) = 0.999$. 
If we use the raw Bradley-Terry logits as our utility, then using that $r(y^1,x) - r(y^0,x) = (r(y^1,x) - r(y^\text{ref},x)) - (r(y^0,x) - r(y^\text{ref},x))$
we have:
\begin{align*}
\frac{\pi^*(y^1 \given x)}{\pi^*(y^0 \given x)} %
  &=\exp(\logit(0.999)-\logit(0.99))) \frac{\pi_0(y^1 \given x)}{\pi_0(y^0 \given x)}  \\
& \approx 10 \times \frac{\pi_0(y^1 \given x)}{\pi_0(y^0 \given x)} 
\end{align*}
That is, when aligning to raw rewards, going from a very good response to a marginally better response increases probability by a factor of 10!
However, if $y^{\text{ref}}$ is already good, a human would find little difference between the two responses.\footnote{Ultimately, the issue here is that Bradley-Terry rewards don't automatically correspond to utilities.}
Conversely, aligning to the transformed reward model only increases the probability by a factor of 1.01.  

This effect seems particularly salient when we consider that the reward model is itself learned. It seems unlikely that the model can actually reliably distinguish between $y_1$ and $y_0$. Accordingly, when we align to the raw learned model, we expect to induce enormous shifts according to preferences that are anyways noisy.

\paragraph{Choosing reference response} 
The reference reward acts as a hyperparameter of the transformation. Essentially, the transformation results in a utility that is linear below the reference value, and rapidly diminishing above it. Accordingly, we should set the reference to a value that represents a good response, that we believe is achievable by the model, and where we believe our learned reward function makes meaningful predictions. We found that a good default choice is the 85th quantile of examples from the base distribution.
We consider additional examples in \cref{sec:experiments}. 

\section{Reward Aggregation}
We now consider the case when we have multiple reward models for different objectives $r_1, ..., r_n$. 

\paragraph{Alignment Goal}
Again, the first step in deriving an optimal aggregation scheme is to formalize a goal.
We make the following natural choice: the aligned model should be ``good'' on all target properties. 
E.g., we want to align our model to be helpful AND harmless.

To formalize this idea, 
let $G_i$ be binary random variable indicating whether $y$ is considered ``good'' for $x$ in property $i$. 
We introduce the binary random variable corresponding to logical AND: $G_\text{AND} := \bigwedge_{i=1}^{n} G_i$. 
Similar to the single reward case, we formalize the goal as the posterior distribution conditioned on all properties being ``good'': 
\begin{align}
    \pi_{\text{AND}, \gamma}^\text{target} \propto \pi_0(y \given x) p(G_\text{AND}=1 \given x, y)^{1/\gamma}
    \label{eq:target_distn_agg}
\end{align}

\paragraph{Reward Aggregation}
With this goal, following \cref{thm:transformation}, we want to align using utility
\begin{align}
    u(x,y) = \log p(G_\text{AND}=1 \given x, y).
\end{align}
The question is then how to construct this utility function using the individual reward models. 

In full generality, this is an impossible task. The reason is that the individual rewards only tell us about the marginal distributions of the properties, but the logical AND may depend on interactions. 
Thus, we need an extra assumption:
\begin{assumption}[Independent Judgements]\label{ass:indep_judge}
Given a fixed prompt $x$ and response $y$, whether $y$ is judged to be good for $x$ on each property is independent of all the judgements on all other properties.
That is, $(G_1,..., G_n )$  are conditionally independent given $(X, Y)$.
\end{assumption}
For example, this assumption says we can decide whether a given response is helpful independently of deciding whether it's harmful. (Note: this is conditional on the prompt and response. We do \emph{not} require helpfulness and harmless to be independent marginally.)

The reward aggregation formula follows immediately:
\begin{gtheorem}
Suppose output $y$ is deemed good for prompt $x$ in aspect $i$ if it would be preferred to reference output $y^\text{ref}_i(x)$ in property $i$.  
Then, if \cref{ass:indep_judge} holds, the Bradley-Terry model \cref{eq:bt} holds for all properties, and we align using KL-regularized utility maximization \cref{eq:rlhf_obj}, then using utility
\begin{align}
  u(x,y) = \sum_{i=1}^n \log \sigma(r_i(x, y) - r^\text{ref}_i(x)) 
    \label{eq:rm_agg}
\end{align}
will satisfy the alignment goal \cref{eq:target_distn_agg}.
Here $r^\text{ref}_i(x) := r(x, y^\text{ref}_i(x))$.
\end{gtheorem}

\paragraph{Mechanistic Interpretation}
We derived the aggregation scheme according to a probabilistic assumption. Similar to the single-reward case, we consider the mechanistic effect.%

The baseline approach is to aggregate with a (weighted) sum of the raw rewards.  The key problem is that this approach allows strong performance in one property to balance out poor performance in another.

Consider the case where we have two properties $A$ and $B$ we want to align to (e.g., helpfulness and harmlessness), and 4 responses $y^\text{ref}_A, y^\text{ref}_B, y^0, y^1$ such that $p(y^\text{ref}_A \prec y^1) = p(y^\text{ref}_B \prec y^1) = 0.9$, and $p(y^\text{ref}_A \prec y^0) = 0.45$ with $p(y^\text{ref}_B \prec y^0) = 0.99$. 
If we want the aligned distribution to generate samples that are ``good'' in both aspects, then $y^1$ should be preferred to $y^0$. 
However, if we use the sum of the raw Bradley-Terry logits as utility ($u = r_A + r_B$), the relative probability ratio under the aligned policy $\pi^*$ will be (with $\gamma=1$)  
\begin{align}
    \frac{\pi^*(y^1 \given x)}{\pi^*(y^0 \given x)} = 1 \times \frac{\pi_0(y^1 \given x)}{\pi_0(y^0 \given x)}.
\end{align} 
That is, the aligned model does not upweight response $y_1$. If we instead align by the sum of the transformed reward, then 
we have that the relative probability ratio is approximately $1.8$---i.e., the response that does well on both properties is preferred.

\section{Experiments}\label{sec:experiments}

\begin{figure}[t]
  \centering
  \begin{subfigure}{\linewidth}
    \begin{subfigure}{.45\linewidth}
      \centering
      \includegraphics[width=\linewidth]{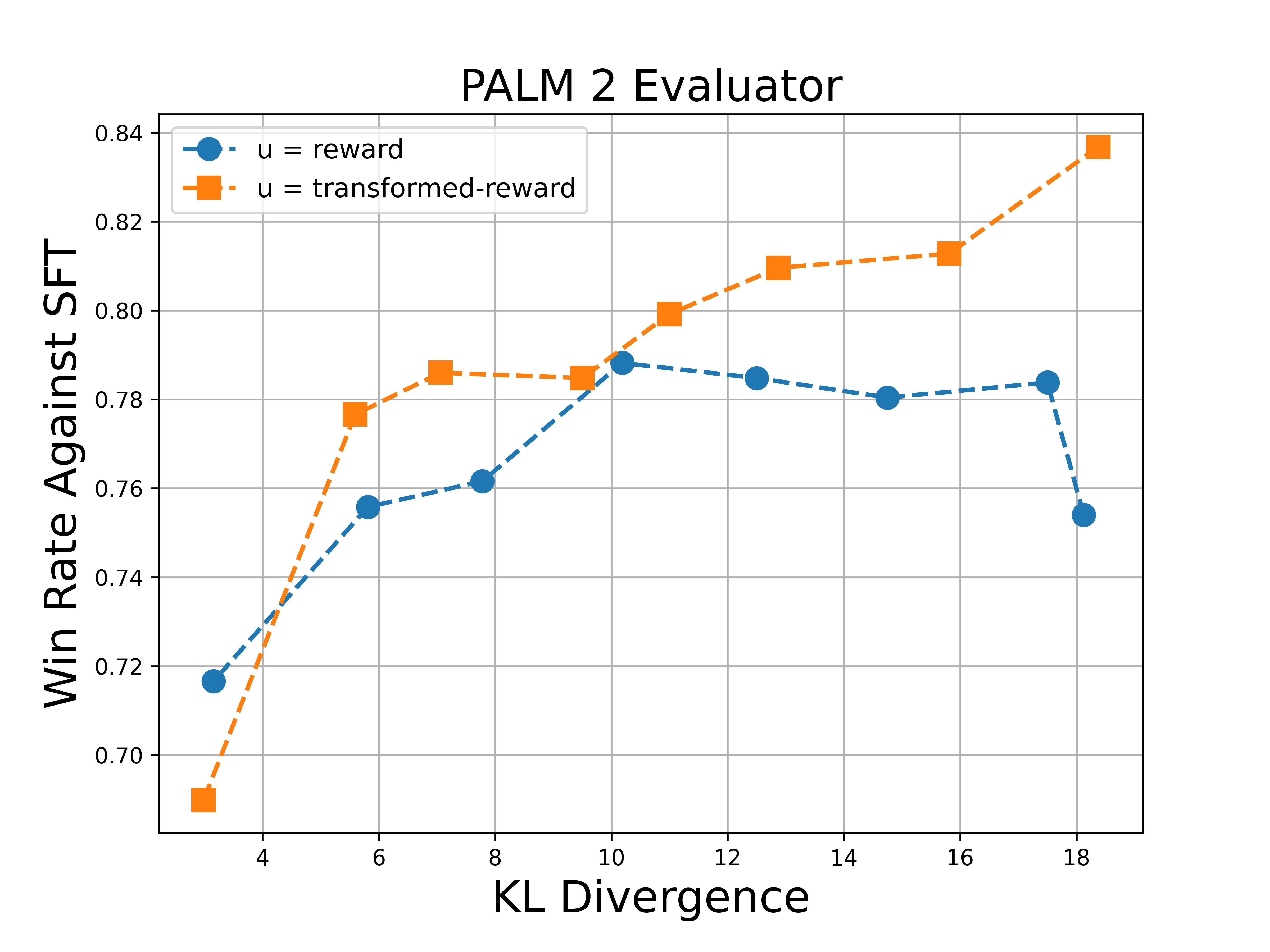}
    \end{subfigure}
    \begin{subfigure}{.45\linewidth}
      \centering
      \includegraphics[width=\linewidth]{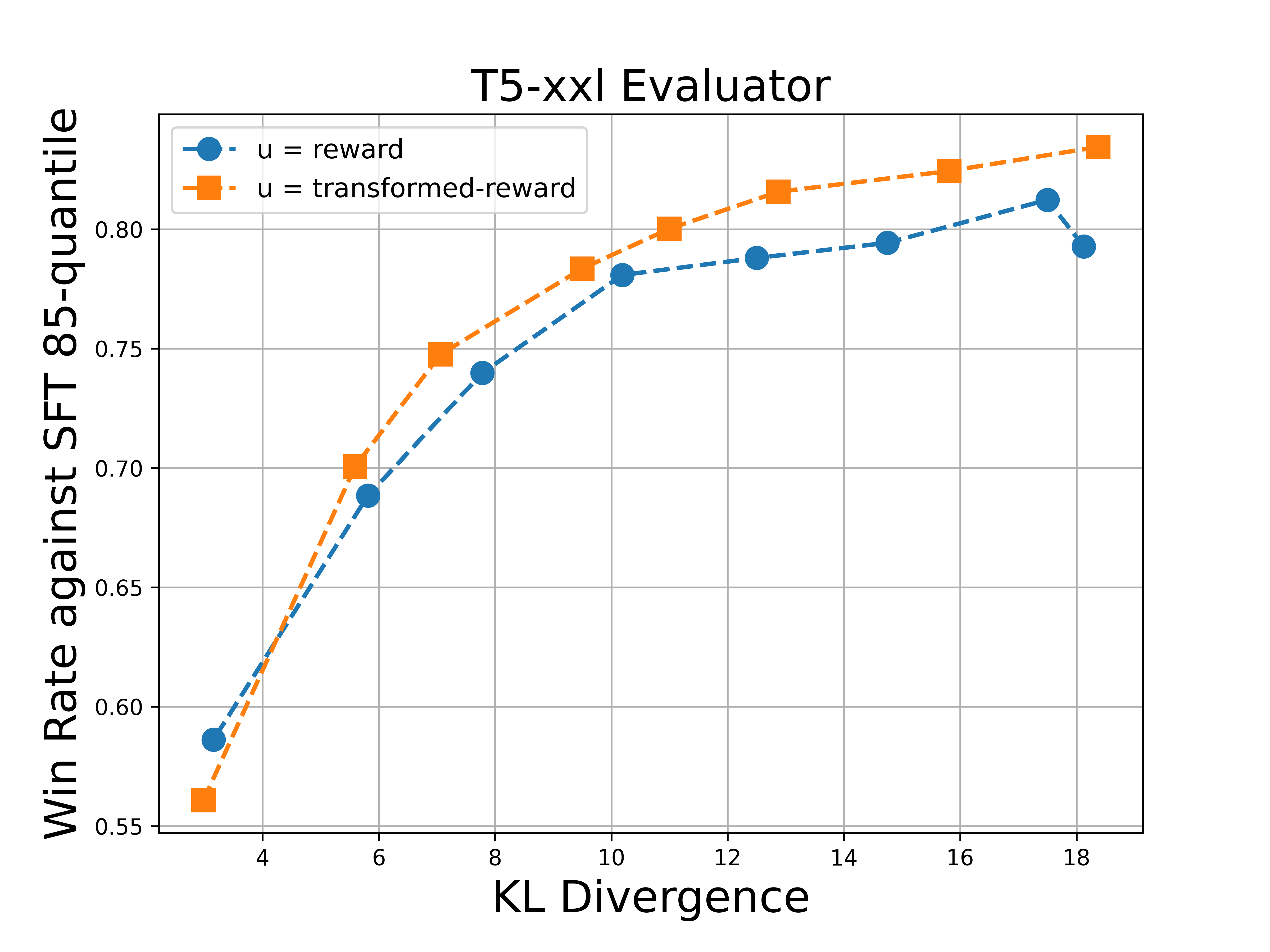}
    \end{subfigure}
    \caption{Helpful: RLHF}
    \label{fig:helpful_wr}
  \end{subfigure}
  
  \vspace{4mm} %

  \centering
  \begin{subfigure}{\linewidth}
    \begin{subfigure}{.45\linewidth}
      \centering
      \includegraphics[width=\linewidth]{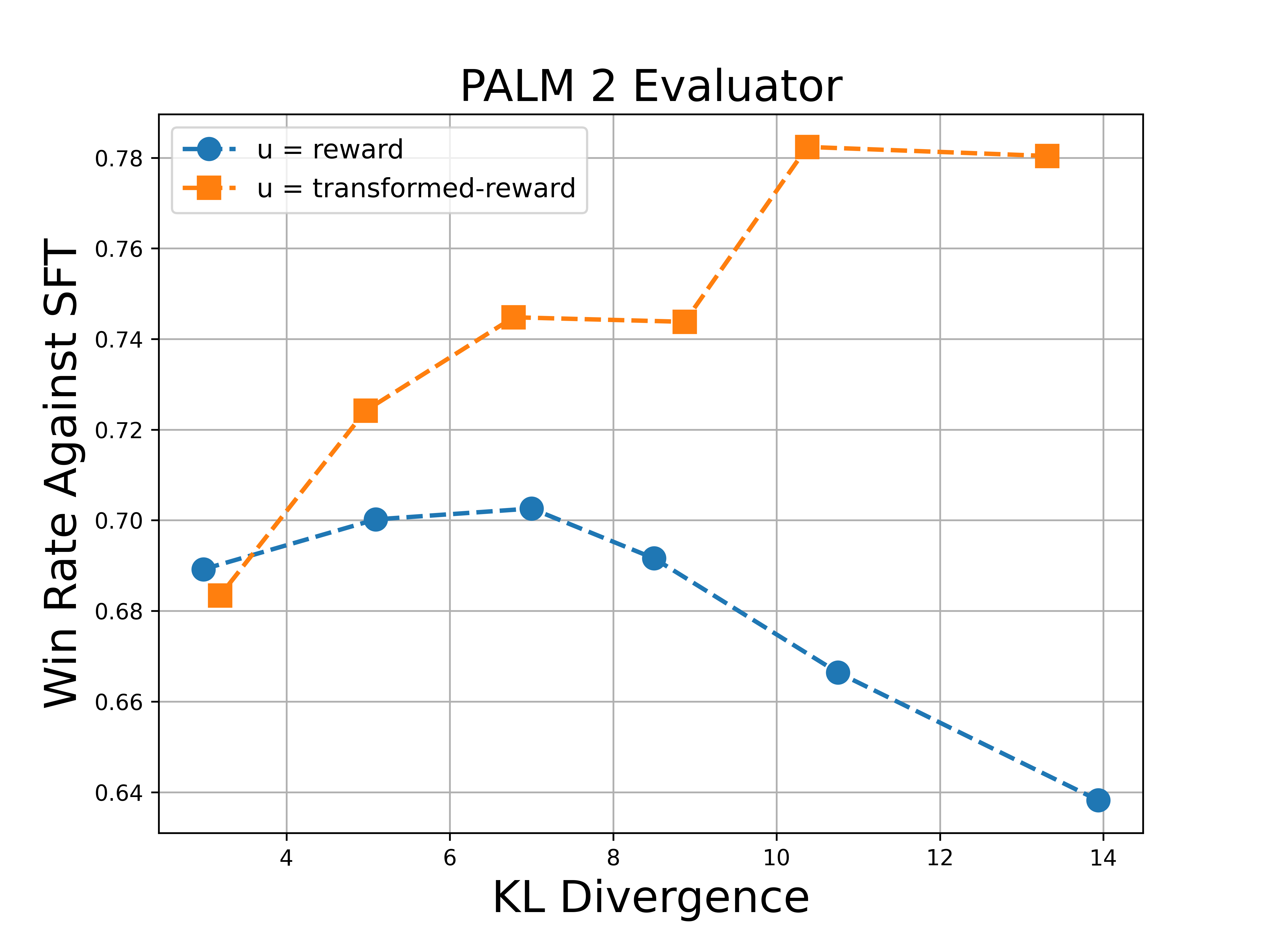}
    \end{subfigure}
    \begin{subfigure}{.45\linewidth}
      \centering
      \includegraphics[width=\linewidth]{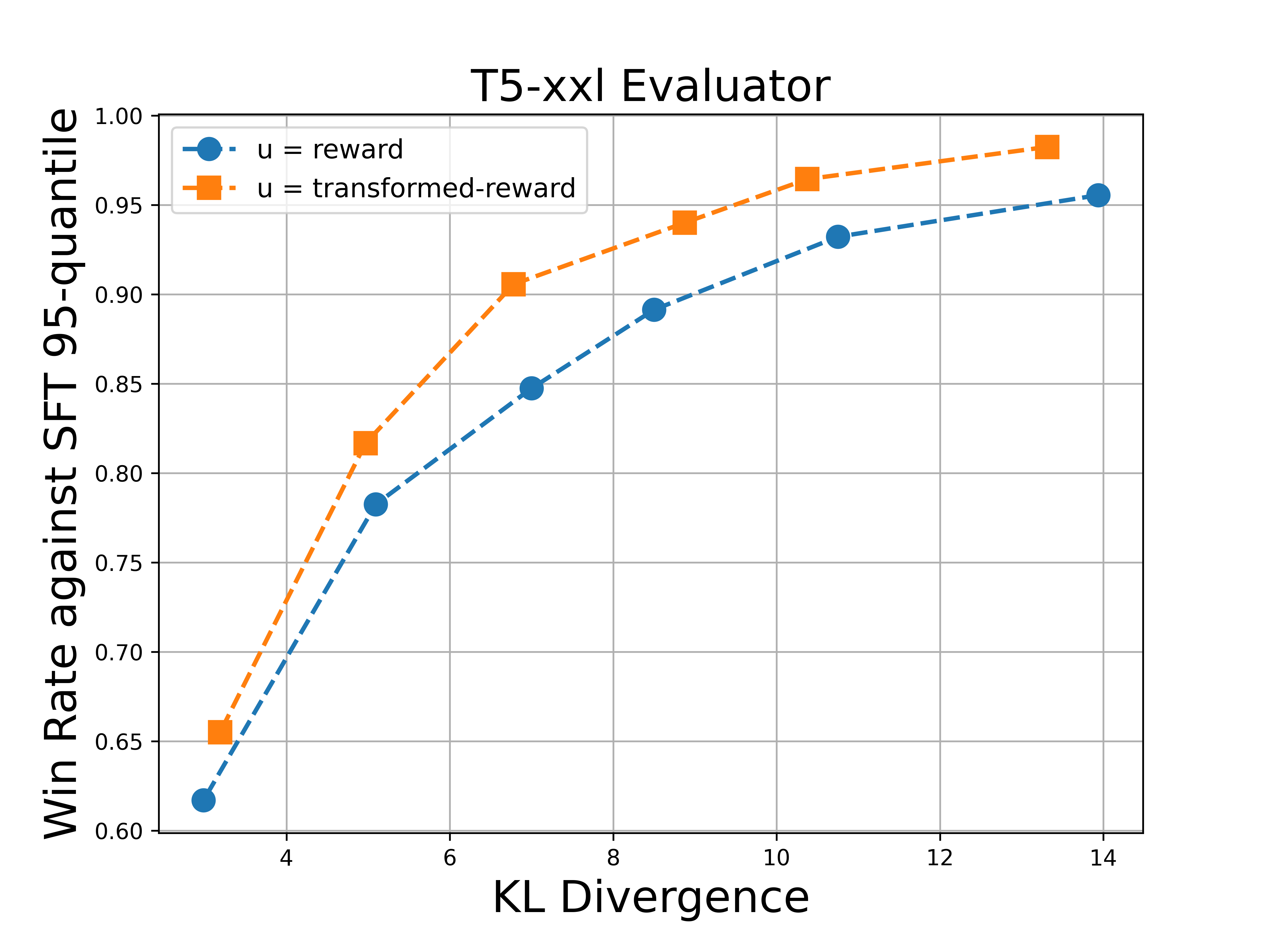}
    \end{subfigure}
    \caption{Harmless: RLHF}
    \label{fig:harmless_wr}  
  \end{subfigure}
  \caption{Transformed reward obtains better KL and win-rate trade-offs in single-reward. We show two win rates:  1) win rates judged from prompted PALM 2 evaluator, between the aligned policy and random SFT samples, and 2) win rates judged by T5-XXL evaluator, against the SFT quantiles: 85\% for helpfulness and 95\% for harmlessness.}
  \label{fig:win_rate_single}
\end{figure}

We now turn to assessing the practical effect of using the transformed reward to align LLMs.
We experiment with aligning models to be helpful, harmless, and both.
We find that the transformation alleviates reward hacking and reward underfitting, and that aligning to transformed sum acts as aligning to logical AND.
This leads to substantial improvements in LLMs aligned to the transformed reward.

\subsection{Experimental Setup}
We follow a standard RLHF pipeline; see \cref{sec:rlhf_details} for full details.
\paragraph{Datasets}
We use the Anthropic Helpfulness and Harmlessness datasets \cite{bai2022training}. 
These are multi-turn dialogues between a human and a digital assistant. Each dataset consists of the beginning of the conversation, two responses for the final turn of the AI side of the conversation, and a label for the human preference on the target property. 
We use the base datasets (44K examples for helpfulness and 42K for harmlessness), where responses are generated from a 52B context-distilled LM. For both tasks, we split the training set into two: half for training the reward model, and half for the alignment step.

\paragraph{Reward model training} We train a Bradley-Terry reward model for each of helpfulness and harmlessness by finetuning a  pretrained T5-base (220M parameters) model \cite{raffel2020exploring} on the Anthropic data. %

\paragraph{SFT} 
For our policy model, we use the instruction-finetuned PALM-2-XXS model~\cite{anil2023PALM}.
Following standard practice, we first run supervised finetuning (SFT) of the instruction tuned LLM on the `preferred' responses from the helpfulness dataset.
We use this SFT model as the pre-alignment base for all experiments.

\paragraph{RLHF setup}
For alignment, we follow standard practice and optimize expected utility subject to a KL penalty using Proximal Policy Optimization (PPO) algorithm. 
For each utility function and dataset, we sweep over multiple values of the KL regularization strength $\gamma$.
We run for $20000$ steps, which we find suffices for convergence in all cases. 

\subsection{LSC-Transformation Improves Alignment}

Transforming the reward model should encourage the alignment to focus on improving lower-reward responses over those with already high rewards. 
We expect this to both reduce reward hacking, and to reduce the number of low-reward responses (less underfitting). 

\paragraph{Choice of reference reward}
The reference reward $r^\text{ref}(x)$ should capture the notion of a response that's ``good enough''.
For harmlessness, this is straightforward: a generic response like ``I can't answer that'' achieves the goal.
For the experiments, we sampled variations of ``I can't answer that'' as the reference reward.

For helpfulness, such canned responses won't suffice.
Instead, for each prompt we sample 64 responses from the SFT model. We then use this data to build an estimator of the 85th quantile of the sampled rewards for each prompt.
We use this estimated 85th quantile as the reference reward. Details provided in \cref{sec:reference_est}.

\paragraph{Transformation Improves Alignment}
Aligning to the transformed reward should reduce reward hacking and underfitting relative to aligning to the raw reward. 
Then, we expect that the transformed alignment leads to larger gains over the SFT model.

We have two main strategies for judging improvement relative to the SFT model.
First, following past work~\cite{gao2023scaling,coste2023reward,eisenstein2023helping}, we train a T5-XXL model using the same preference dataset and the Bradley-Terry objective.
This provides a proxy for true preferences that we do not optimize against (so, it can witness reward hacking).
We say a sample from the aligned model wins in helpfulness if it beats the 85th reward quantile of samples from the SFT model.
We say it wins in harmlessness if it beats the 95th-quantile.
(These numbers are chosen to make winning hard enough to show a gap between alignment strategies; results are consistent across other choices of quantile.).

The second strategy evaluates wins by zero-shot querying of an instruction-tuned PALM-2 medium model.
Following previous work \cite{dubois2023alpacafarm, singhal2023long, eisenstein2023helping, rafailov2023direct},
we pass a prompt, a response from the SFT model, and a response from the aligned model and ask which is preferred (in terms of helpfulness or harmlessness). Details in \cref{sec:PALM2_eval}.

\Cref{fig:win_rate_single} %
 shows the win rate of the aligned models over the SFT model for each evaluation strategy.
We average over the prompts in the RLHF validation dataset.
We see that aligning to the transformed reward dominates aligning to the raw reward, under both evaluation strategies and at all levels of KL distance to the base policy model. 

See \cref{sec:PALM2_eval} for additional experiments evaluating alignment-induced improvements.

\begin{figure}[t]
  \centering
  \begin{subfigure}[b]{0.49\columnwidth}
    \includegraphics[width=\linewidth]{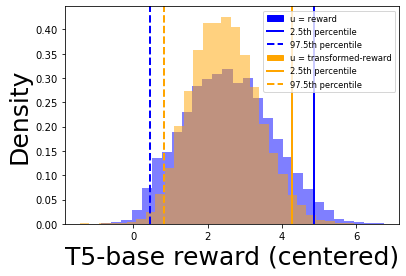}
    \caption{Helpfulness ($\text{KL} \approx 18$)}
  \end{subfigure}%
  \begin{subfigure}[b]{0.49\columnwidth}
    \includegraphics[width=\linewidth]{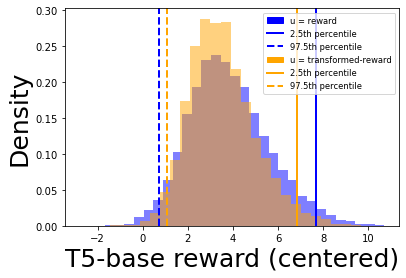}
    \caption{Harmlessness ($\text{KL} \approx 7$)}
  \end{subfigure}
  \captionsetup{skip=1pt}
  \caption{Reward Transformation leads to more uniform reward improvements than baseline. We compare reward distributions in the aligned policies that are matched on KL. Rewards are centered by the SFT median in both helpfulness and harmlessness. Reward distributions are more concentrated when using transformed rewards than using raw rewards.}
  \label{fig:reward_distn}
\end{figure}

\paragraph{Uniform Improvement of Rewards}
It is clear that aligning using the transformed reward improves over aligning using the raw reward.
Intuitively, this is because of a reduction in both reward hacking and underfitting.
To check whether this intuition holds, 
we plot the distribution of rewards of samples from (approximately) KL-matched raw-aligned and transformed-aligned models in \cref{fig:reward_distn}.
As expected, the reward distribution of samples from the transformed-aligned model is more concentrated.
That is, there are fewer very high reward samples (less reward hacking) 
and fewer very low reward samples (less underfitting).

In more detail: for each of helpfulness and harmlessness, we choose a raw-aligned and transformed-aligned model with approximately matched KL. 
We sample responses from each model with the same prompts. 
Then, we compute the (T5-base) reward for each of these responses, centered by median reward of SFT samples (to make rewards comparable across prompts). 
\Cref{fig:reward_distn} shows histograms of these sampled rewards. We also compare across multiple KL-values in \cref{fig:quantile_vs_kl}.  

\paragraph{Transformation reduces shortcuts}

\begin{figure}[t]  %
  \centering
  \begin{subfigure}{0.49\columnwidth}  %
    \centering
    \includegraphics[width=\linewidth]{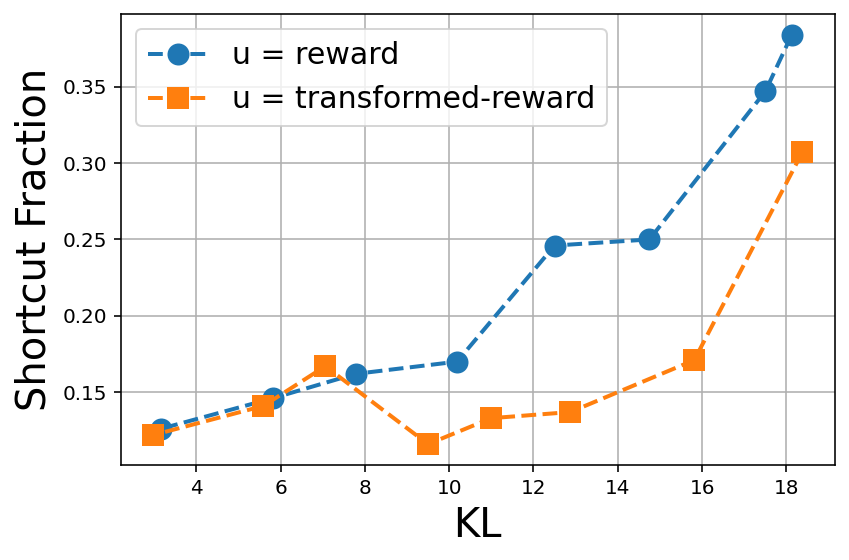}
    \caption{Helpfulness}
    \label{fig:help_shortcut}
  \end{subfigure}%
  \begin{subfigure}{0.49\columnwidth}  %
    \centering
    \includegraphics[width=\linewidth]{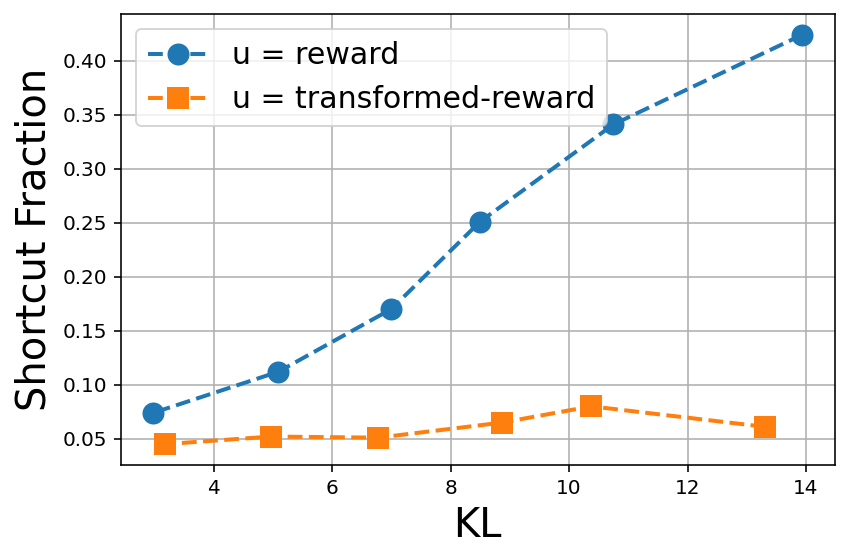}
    \caption{Harmlessness}
    \label{fig:harm_shortcut}
  \end{subfigure}
  \caption{Transformed reward reduces shortcuts in generated responses. For helpfulness, we identify a shortcut pattern of using lists, similar to \cite{eisenstein2023helping}. In harmlessness, one known shortcut pattern is recommending the users to seek therapy or consult professional help \cite{bai2022training}. We extract these shortcuts with heuristic methods. In the baseline approach, the policy model exploits those shortcuts for higher reward values. This is mitigated when we transform the reward.}
  \label{fig:shortcuts}
\end{figure}
 One symptom of reward hacking is that the aligned model will start exploiting ``shortcuts'' that are preferred by the reward model (but which do not correspond to genuine improvements).
 We consider the effect of reward transformation on two such shortcuts.
 For helpfulness, \citet{eisenstein2023helping} observe that raw-aligned models have a tendency to format outputs as lists.
 For harmlessness, we observe that raw-aligned models will often give responses of the form ``you should consult a [doctor/psychologist/lawyer/etc]'' (a similar observation is made by \citet{bai2022training}). 
 In \cref{fig:shortcuts}, we plot the fraction of responses that contain each shortcut, for each aligned model.
 We see that the raw-reward aligned model does indeed exploit these shortcuts. Further, this behavior is substantially mitigated by aligning to the transformed reward instead. See \cref{sec:shortcut_heuristics} for details.
 
 \subsection{Reward Aggregation using LSC-Transformation Significantly Improves Alignment}
 \begin{figure}[h]
    \begin{subfigure}{0.45\linewidth}
        \centering
        \includegraphics[width=\linewidth]{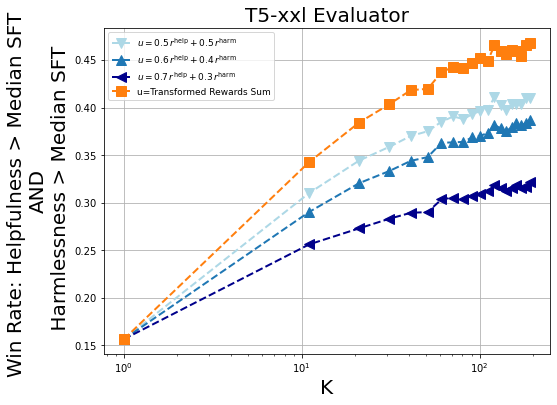}
        \caption{Helpful and Harmless: Best-of-$k$}
        \label{fig:bon_xxl_wr_agg}
    \end{subfigure}
    \hspace{0.5cm}
    \begin{subfigure}{0.45\linewidth}
        \centering
        \includegraphics[width=\linewidth]{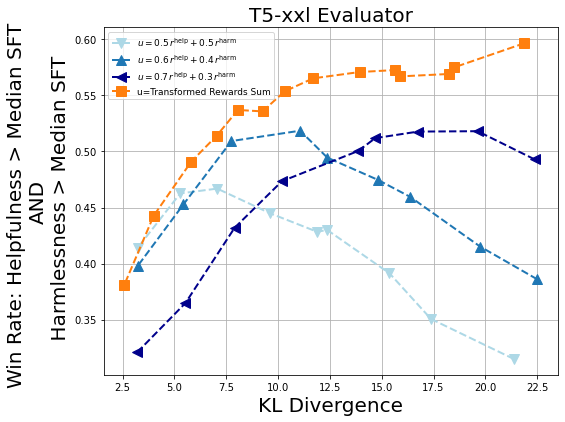}
        \caption{Helpful and Harmless: RLHF}
        \label{fig:rlhf_xxl_wr_agg}
    \end{subfigure}
    \caption{Summation of transformed reward obtains better trade-off between KL/$K$ and win-rate. In \cref{fig:bon_xxl_wr_agg} and \cref{fig:rlhf_xxl_wr_agg} we show win rates against SFT median rewards in both helpfulness and harmlessness, judged by T5-XXL evaluator, for Best-of-$k$ and RLHF respectively.}
    \label{fig:aggregation_wr}
\end{figure}
We now turn to the second goal: aggregating rewards for multiple distinct goals.
To that end, we consider aligning a LLM to be both helpful and harmless.

\subsubsection{Further Experimental Setup}
\paragraph{RLHF setup}
We use reward models trained for the helpfulness and harmlessness tasks as discussed above.
For RLHF training, we use prompts only from the helpfulness dataset. This decision is because of the observation of the tension between helpfulness and harmlessness, which forced \cite{bai2022training} to use a higher proportion of helpfulness prompts than harmlessness ones.  
We use the same policy model as in experiments for single rewards (SFT-ed on helpfulness data). The other training details are the same as in single-reward experiments. 
\paragraph{best-of-$k$ setup}
In addition to RLHF experiments, we also experiment with best-of-$k$ sampling, as in \Citet{gao2023scaling, eisenstein2023helping}.
That is, we draw $k$ samples from the SFT model, rank them by the combined reward, and return the top ranked sample. 
This can be viewed as another alignment procedure, where the best-of-$k$ sampling has some (KL) divergence from the underlying policy, and produces samples with higher expected reward.
In our experiments, we try $k$ increasing from 1 to 191. 

There are two main motivations for considering best-of-$k$ experiments.
First, it demonstrates that the reward aggregation method applies to methods beyond RLHF. %
Second, best-of-$k$ doesn't involve complicated optimizations. This allows us to disentangle the effect of having the `right' reward from the effect on solving the optimization problem. 

\paragraph{Baseline}
In this setting, we take the baseline method to be a weighted sum of the raw rewards; i.e.,
\begin{align}
    R^\text{baseline}_\land \defeq w R^\text{help} + (1-w)R^\text{harm}.
\end{align}
We sweep over $w=0.1,..., 0.9$, and report the baselines with best performance on our evaluation metrics. 

\paragraph{Reference Rewards}
We combine the transformed rewards by simple addition. 
Weighting this addition doesn't have a clear motivation or interpretation. 
Instead, the choice of reference value used for each reward plays an analogous role. Intuitively, if we set a lower reference value for reward A then the reward becomes more easily saturated, and the optimization focuses more on reward B.

For best-of-$k$, using $w=0.5$ achieves the best performance in the baseline method. Accordingly, we want to take the reference rewards for helpfulness and harmlessness to be comparable. To that end, we use the (estimated) SFT median for both helpfulness and harmlessness. Note that this is a lower reference than used for the individual optimizations. This makes sense because the more difficult problem of optimizing both goals simultaneously stops the model from saturating the reward prematurely.

For RLHF, we observe that $w=0.6$ gives the best result for weighted-sum approach. Then, we want to set the reference reward for helpfulness to be somewhat higher. We use the (estimated) SFT 75\%-quantile.

These choices are likely not optimal, and it's possible that further hyperparameter optimization could improve results.

\paragraph{Evaluation}
We want a metric that reflects Logical-AND. That is, whether we've improved over the SFT model in both helpfulness and harmlessness. To that end, we'll say that a generation wins if it has helpfulness reward higher than the median helpfulness of SFT samples, and harmlessness higher than the median harmlessness of SFT samples. 

\subsubsection{Aggregation Results}

\begin{figure}[t]
  \centering
  \begin{subfigure}{\linewidth}
    \centering
    \includegraphics[width=\linewidth]{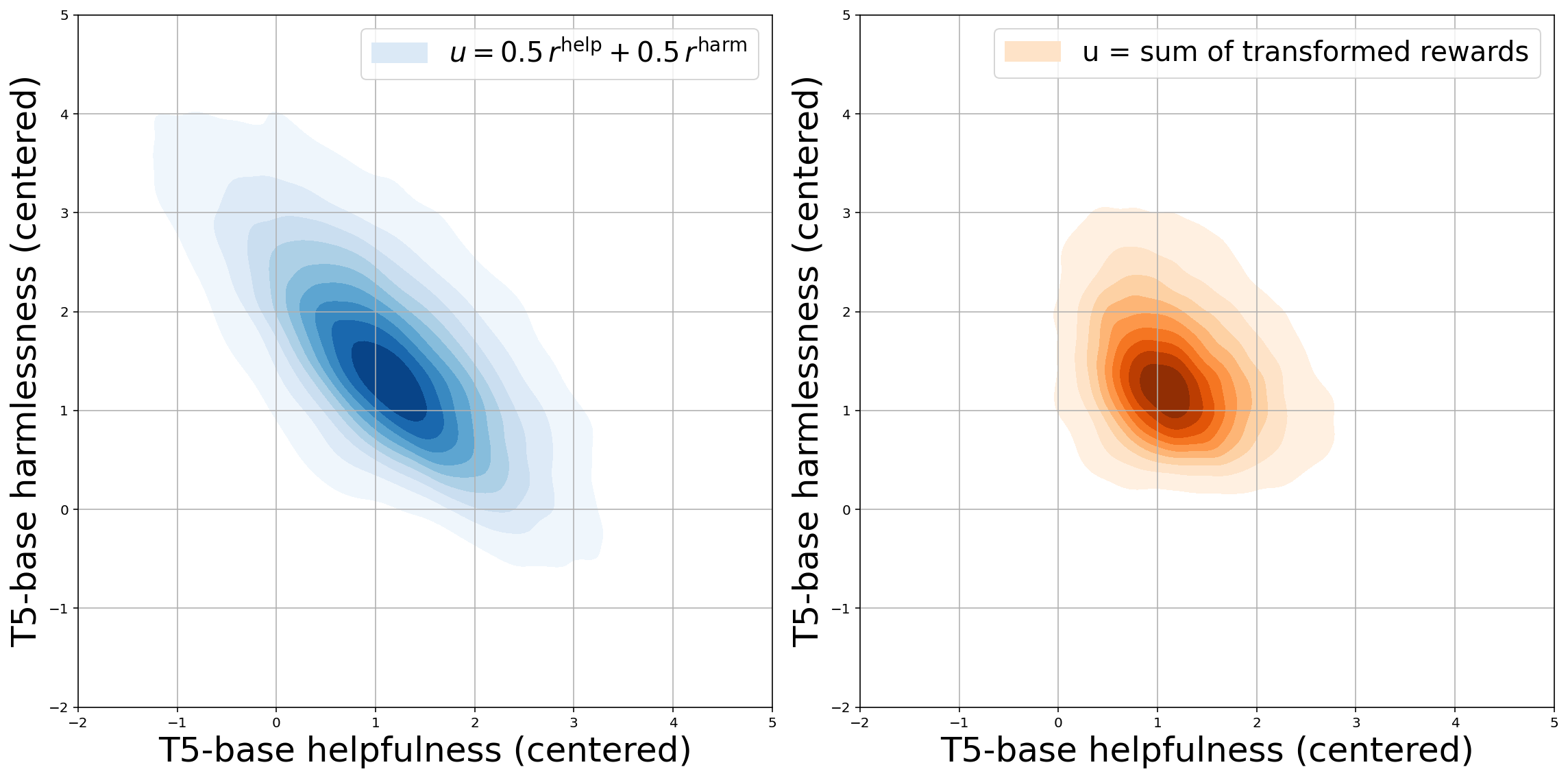}
    \caption{best-of-$k$ ($K=191$)}
    \label{fig:bon_kde}
  \end{subfigure}

  \begin{subfigure}{\linewidth}
    \centering
    \includegraphics[width=\linewidth]{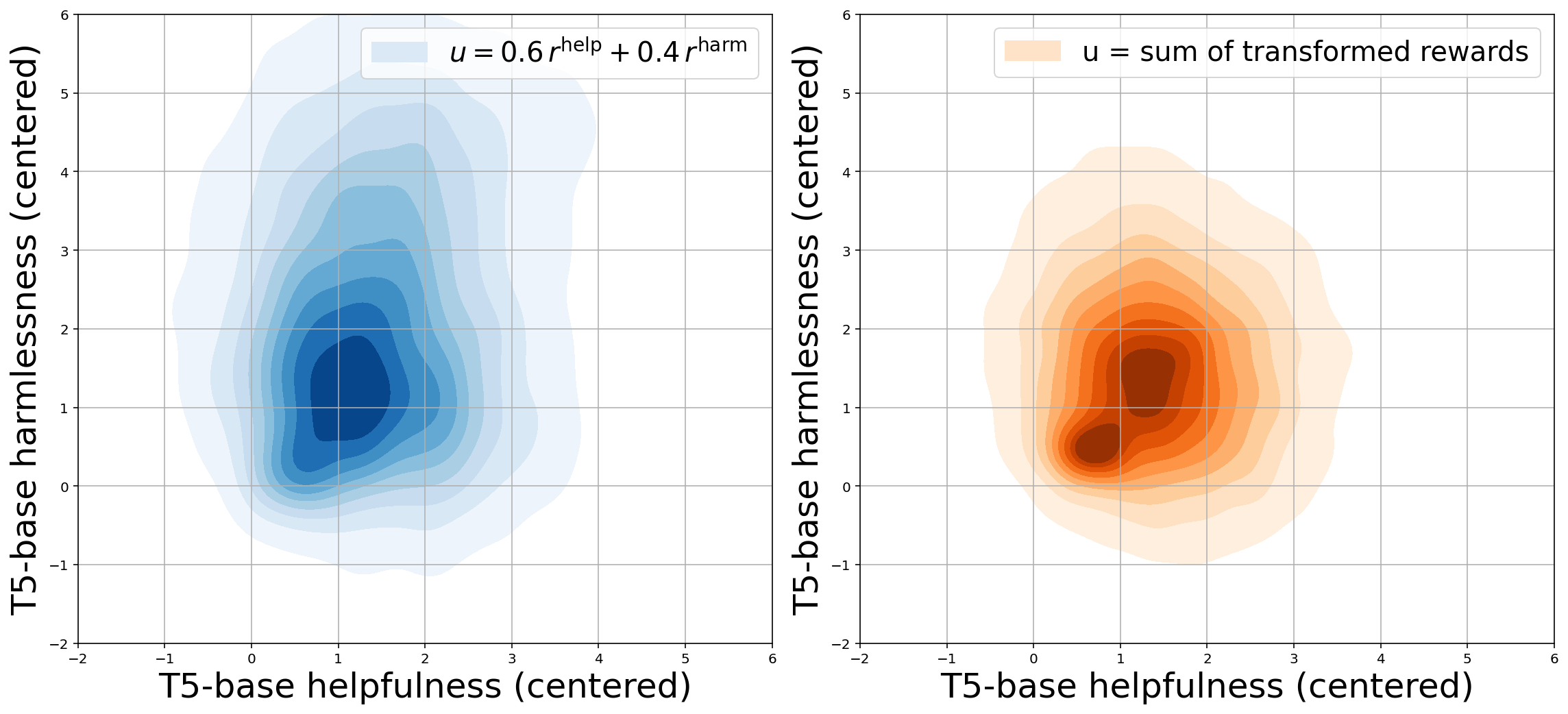}
    \caption{RLHF ($\text{KL} \approx 14$)}
    \label{fig:rlhf_kde}
  \end{subfigure}
  \caption{Summation of log-sigmoid transformed rewards corresponds better to logical AND. Aligned policies using the former method have more balanced reward distributions (concentrated where the two reward values are similar), whereas the latter method leads to more unbalanced reward distributions. We choose aligned policies by matching $K$ for best-of-$k$ and $\text{KL}$ for RLHF. The rewards are centered by SFT median for both helpfulness and harmlessness. We visualize the joint distribution by kernel density estimate plot, with darker color indicating higher density.}
  \label{fig:kde}
\end{figure}
\paragraph{Transformed Aggregation Improves Alignment}
Summing the transformed reward should have two advantages over the baseline method.
First, it corresponds to logical AND. Second, it retains the benefits of alleviating reward overoptimization, as in the single-reward case. 
Together, these should cause aligning by the transformed-combined reward to outperform aligning by the baseline reward.

\Cref{fig:bon_xxl_wr_agg} and \Cref{fig:rlhf_xxl_wr_agg} shows improvement over the SFT model for aligning using both rewards, for both best-of-$k$ and RLHF. 
As anticipated, we see significant improvement from the transformed method.

It is noteworthy that the transformed-aligned model outperforms in best-of-$k$, and for low-KL in the RLHF alignment.
In these cases, there is little reward hacking (the aligned model is too close to the SFT model to have very bad behavior). 
Thus, the win here is apparently due mainly to the logical AND effect.
In the high-KL regime, the reward hacking effect kicks in and the transformed-reward dramatically outperforms the raw-reward baseline.

\paragraph{Transformed Summation Corresponds to Logical AND}
Next, we check directly whether the logical AND effect can be witnessed in the aligned LLM. 
To that end, we examine the distributions of rewards in the aligned policies.
In \cref{fig:kde}, our aggregation method leads to more balanced reward distributions (two reward values are often similar), whereas the baseline method leads to more unbalanced reward distributions (one reward is much higher than the other one). Note that best-of-$k$ ($k=191$) have quite different reward distributions than RLHF at $\text{KL} \approx 14$ (the former reflects the anti-correlation of the two rewards in the initial policy; the latter updates the policy to have high aggregated reward values). But the transformed aggregation method has consistent effects in both cases.

\section{Ablation Studies}
\begin{figure}[ht]
  \centering
  \includegraphics[width=\linewidth]{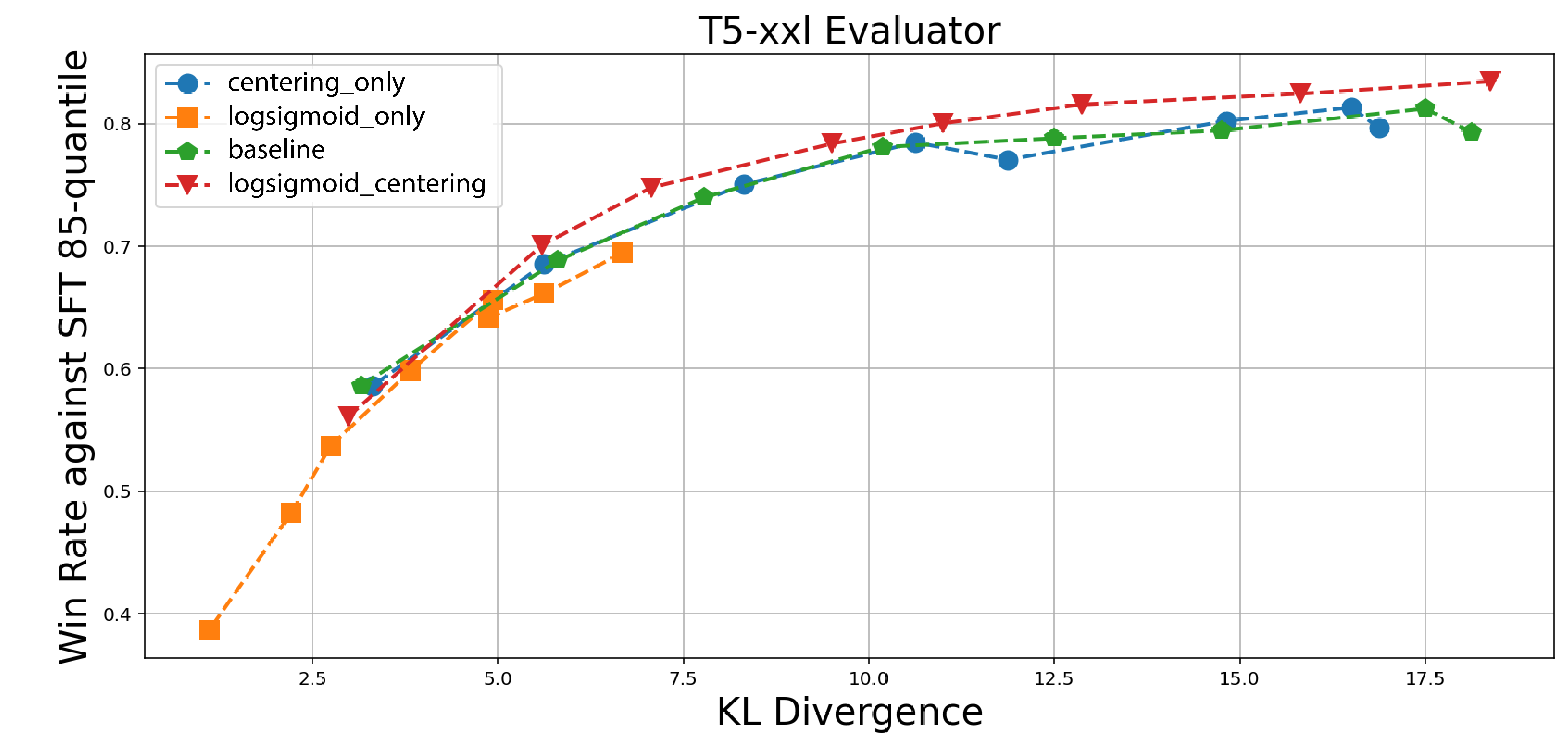}
  \vspace{-16pt}
  \caption{Log-sigmoid transformation or centering alone does not improve alignment. The experiment is for the helpfulness task.}
  \label{fig:ablation}
  \vspace{-3mm}
  \end{figure}

The LSC-transformation consists of two components: centering and log-sigmoid transformation. Centering is known to reduce variance in policy gradient updates, while the log-sigmoid transformation caps the reward values. To understand the contribution of each component, we investigate their effects separately. As shown in \cref{fig:ablation}, applying centering or log-sigmoid transformation in isolation does not improve alignment.

\paragraph*{Does centering alone help?}
This approach did not change the RLHF performance relative to the baseline. Though centering is known to reduce variance for policy gradient update, it doesn’t change the optimization target of the KL-regularized RLHF objective, as shown in \cref{eq:aligned_dist}.

\paragraph*{Does log-sigmoid transformation alone help?}
This essentially sets a constant reference value ($r_\text{ref}(x) = 0$) for all inputs $x$. This is found to hurt the performance relative to baseline. The reference value is important, as can be seen in the shape of utility function in Fig 2(a): when the reference is too small, it saturates the utility when the reward is still under-fitted; when the reference is too large, the nonlinear transformation has no effect. More specifically, there are two possible reasons why simply using log-sigmoid transformation does not work:
\begin{enumerate}
  \item First, $r(x, y)$ is non-identifiable: any $r(x, y) + c(x)$ yields the same Bradley-Terry objective. Then, any fixed constant reference reward (independent of prompt) is fundamentally not meaningful.
  \item Second, if the reference reward is set too high then there will be no effect (for $r \ll r_\text{ref}$, the transformation is effectively a constant offset), and if the reference reward is set too low then training will saturate too early. The issue with a constant reference level such as 0 is that it’s unclear what this means in practical terms—is it too high or too low or fine? By contrast, choosing reference rewards based on generated responses (as in the paper) makes it easy to set a meaningful threshold.
\end{enumerate}

\section{Discussion and Related Work}
There is a growing body of work on mitigating reward hacking in the RLHF pipeline.
Techniques include forms of reward model averaging \cite{eisenstein2023helping, rame2024warm, zhai2023uncertainty}, constrained optimization \cite{moskovitz2023confronting}, and reward model regularization \cite{shen2023trickle}, iterative human preference collection \cite{bai2022training, stiennon2020learning, fan2022nano}, or data bias mitigation \cite{singhal2023long}.
 These approaches are complementary to the transformation technique proposed here, and could be used in combination. 

There have been several proposals for aligning language models to multiple objectives. %
The most common approach is to combine individual reward models via a weighted sum \citep[e.g.,][]{wu2023finegrained, moskovitz2023confronting}.
\Citet{moskovitz2023confronting} identified a constraint threshold for individual rewards, and formalized a constrained MDP problem, but the identification of the threshold point relies on ground-truth queries.   
\Citet{bakker2022finetuning} consider adapting social welfare schemes for aggregating the dissenting preferences of many individuals, in contrast with our goal of satisfying all properties. 
\Citet{bai2022training} train a single reward model on both the helpfulness and harmlessness data, but discover that this leads to reward hacking harmlessness. They change the proportion of helpfulness data in the training to circumvent this. Combining by summing transformed rewards allows us to circumvent such considerations.

The transformation technique in this paper is relevant to any alignment strategy that explicitly maximizes an expected utility.
There are now also alignment methods \citep[e.g.,][]{rafailov2023direct, azar2023general,zhao2022calibrating} that use preference labels directly without explicitly instantiating reward models.
Note, however, that if we want to align to multiple properties, we still need to compute rankings from an aggregate. The simplest approach is to train reward models for individual properties, combine these reward models, and then rank samples using the combine reward. Our best-of-k experiments show that the transformation can yield significant gains even in this case.

Finally, we note that 
\citet{azar2023general} also emphasizes the need for a bounded utility function.
The work here can be viewed, in part, as a way of incorporating this insight that still maintains the standard utility maximization pipeline. It is an interesting question for future work whether making explicit use of a (transformed) reward improves the alignment step relative to using only ranked pairs.

\clearpage
\section*{Acknowledgements}
This work is supported by ONR grant N00014-23-1-2591 and Open Philanthropy.
\section*{Impact Statement}
This paper presents work whose goal is to advance the field of Machine Learning. There are many potential societal consequences of our work, none which we feel must be specifically highlighted here.

\printbibliography

\newpage
\appendix

\section{Additional Experiment Details}\label{sec:additional_results}
\subsection{Implementation for reference value prediction}\label{sec:reference_est}
The goal is to develop models that can predict various quantiles of the sampled rewards for different prompts.
Instead of building separate models for each quantile, we assume the reward distribution for each prompt follows a Gaussian distribution.
Under this assumption, we build two models: \textbf{Mean Prediction Model} (\(r^{\text{mean}}\)) and \textbf{Standard Deviation Prediction Model} (\(r^{\text{std}}\)) to predict the mean and standard deviation of the reward distribution for each prompt $x$.

To estimate a specific quantile for a prompt, the formula used is:
\[ r^{\text{mean}}(x) + s \times r^{\text{std}}(x) \]
where \( s \) is a scaling factor corresponding to the desired quantile (e.g., using \( s=0.7 \) approximately gives us the 75\% quantile).

The training data is collected by generating 64 responses per prompt using the SFT policy model. These responses are scored to calculate the sample mean and standard deviation of the reward distribution for each prompt, which are then used as the response variables in training the models \(r^{\text{mean}}\) and \(r^{\text{std}}\).

\subsection{PALM-2 Evaluation Details}\label{sec:PALM2_eval}
\begin{figure}
    \centering
    \includegraphics[width=\columnwidth]{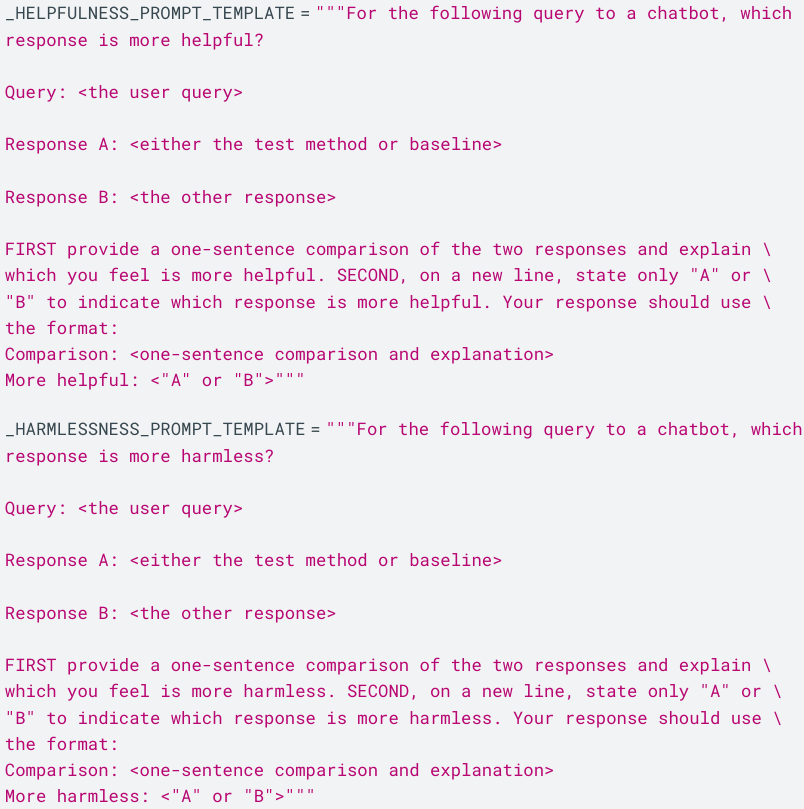}
    \caption{Prompt for PALM-2 Evaluation}
    \label{fig:PALM2_eval}
\end{figure}

We evaluate the win-rates with zero-shot prompting. 
For each prompt, we sample $(y_{\pi}, y_{\text{sft}})$ from the aligned and SFT policy, then ask the PALM-2 model which is more helpful/harmless. 
The prompt template is \cref{fig:PALM2_eval}.  

In order to counter positional bias \cite{hou2023large}, we run PALM-2 on the two possible orderings $(y_{\pi}, y_{\text{sft}})$ and $(y_{\text{sft}}, y_{\pi})$, sample $N = 8$ outputs for each order and determine the winner by majority voting.

\subsection{Heuristics for shortcut discovery}\label{sec:shortcut_heuristics}
For the helpfulness task, we follow the same practice in \cite{eisenstein2023helping} and find responses in the format of a list. 

For harmlessness task, we see the aligned policy starts to only generate similar generations to different questions as KL gets larger. 
To find shortcuts, we analyze generations of the aligned policy (using raw reward as utility) at $\text{KL} \approx 14$. 
Then we perform topic modeling and identify one dominant topic characterized by the recurring theme of "you should talk to [profession]."  
From the most frequent 50 words in that topic, we find all the profession-related words: therapist, professional, doctor, psychologist, counselor, lawyer, police. 
Then we find all responses containing profession-related words. 

\subsection{RLHF training details}
\label{sec:rlhf_details}
We use Proximal Policy Optimization (PPO) to perform RLHF alignment. The specific hyper-parameters are in \cref{tab:hyper_param}
\begin{table}[ht]
\centering
\caption{Hyper-parameters for RLHF.}
\begin{tabular}{@{}ll@{}}
\toprule
Parameter                   & Value          \\ \midrule
Policy learning rate        & $5 \cdot 10^{-6}$ \\
Value learning rate         & $4 \cdot 10^{-5}$       \\
Learning schedule           & Constant (linear warm-up) \\
Training steps              & 20000           \\
Warm-up steps               & 2000            \\
Batch size                  & 32         \\
Input length                & 1024            \\
Output length               & 256             \\
 \bottomrule
\end{tabular}
\label{tab:hyper_param}
\end{table}
We sweep over $\gamma$'s to get aligned policies with different KL values. Since we want to match (converged) polices by their KL values, we find some heuristic to predict converged KL for a chosen $\gamma$ (so they may not look like regular linear or geometric series). More specifically, we use the parameterization of $\alpha = \frac{\gamma}{1 + \gamma}$, and the values used are in \cref{tab:alpha}. Note that to get the same KL value, we use smaller KL regularization for reward transformation, than using the raw reward. This is intuitive as the log-sigmoid transformation prevents using KL budgets once the reward saturates. 

\begin{table*}[ht]
\centering
\caption{Alphas for Various Task-Methods.}
{\scriptsize %
\begin{tabular}{@{}ll@{}}
\toprule
\textbf{Task-Method} & $\bm{\alpha := \frac{\gamma}{1+\gamma}}$\\ 
\midrule
Helpfulness (u = reward) & [0.081, 0.086, 0.092, 0.1, 0.114, 0.132, 0.161, 0.222] \\
Helpfulness (u = transformed reward) & [0.02, 0.023, 0.028, 0.032, 0.039, 0.053, 0.075, 0.123] \\
Harmlessness (u = reward) & [0.169, 0.182, 0.196, 0.218, 0.248, 0.32] \\
Harmlessness (u = transformed reward) & [0.032, 0.041, 0.052, 0.079, 0.126, 0.222] \\
H+H ($0.5 r^\text{help} + 0.5 r^\text{harmless}$) & [0.078, 0.08, 0.084, 0.088, 0.094, 0.102, 0.116, 0.134, 0.168] \\
H+H ($0.6 r^\text{help} + 0.4 r^\text{harmless}$) & [0.068, 0.071, 0.074, 0.077, 0.082, 0.088, 0.1, 0.123, 0.163] \\
H+H ($0.7 r^\text{help} + 0.3 r^\text{harmless}$) & [0.057, 0.06, 0.065, 0.07, 0.075, 0.084, 0.1, 0.126, 0.173] \\
H+H (sum of transformed reward) & [0.014, 0.0145, 0.015, 0.017, 0.018, 0.02, 0.023, 0.026, 0.031, 0.034, 0.04, 0.048, 0.066, 0.096] \\
\bottomrule
\end{tabular}
}
\label{tab:alpha}
\end{table*}

\section{More Experiment Results}
\begin{figure*}
    \centering
    \begin{subfigure}{0.48\textwidth}
        \centering
        \includegraphics[width=\linewidth]{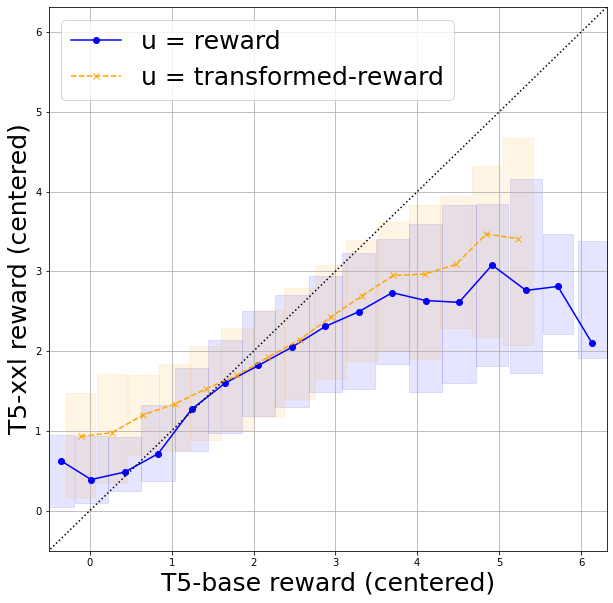}
        \caption{Helpful (single reward): $\text{KL} \approx 18$}
        \label{fig:help_base_vs_xxl_single}
    \end{subfigure}
    \begin{subfigure}{0.48\textwidth}
        \centering
        \includegraphics[width=\linewidth]{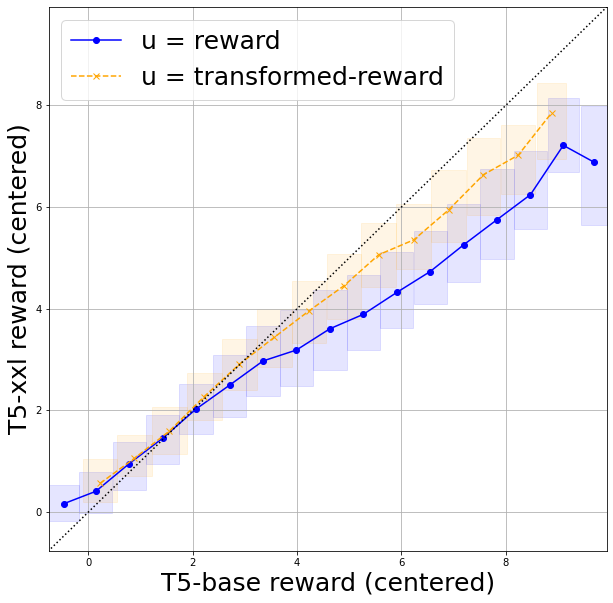}
        \caption{Harmless (single reward): $\text{KL} \approx 7$}
        \label{fig:harm_base_vs_xxl_single}
    \end{subfigure}
    
    \begin{subfigure}{0.48\textwidth}
        \centering
        \includegraphics[width=\linewidth]{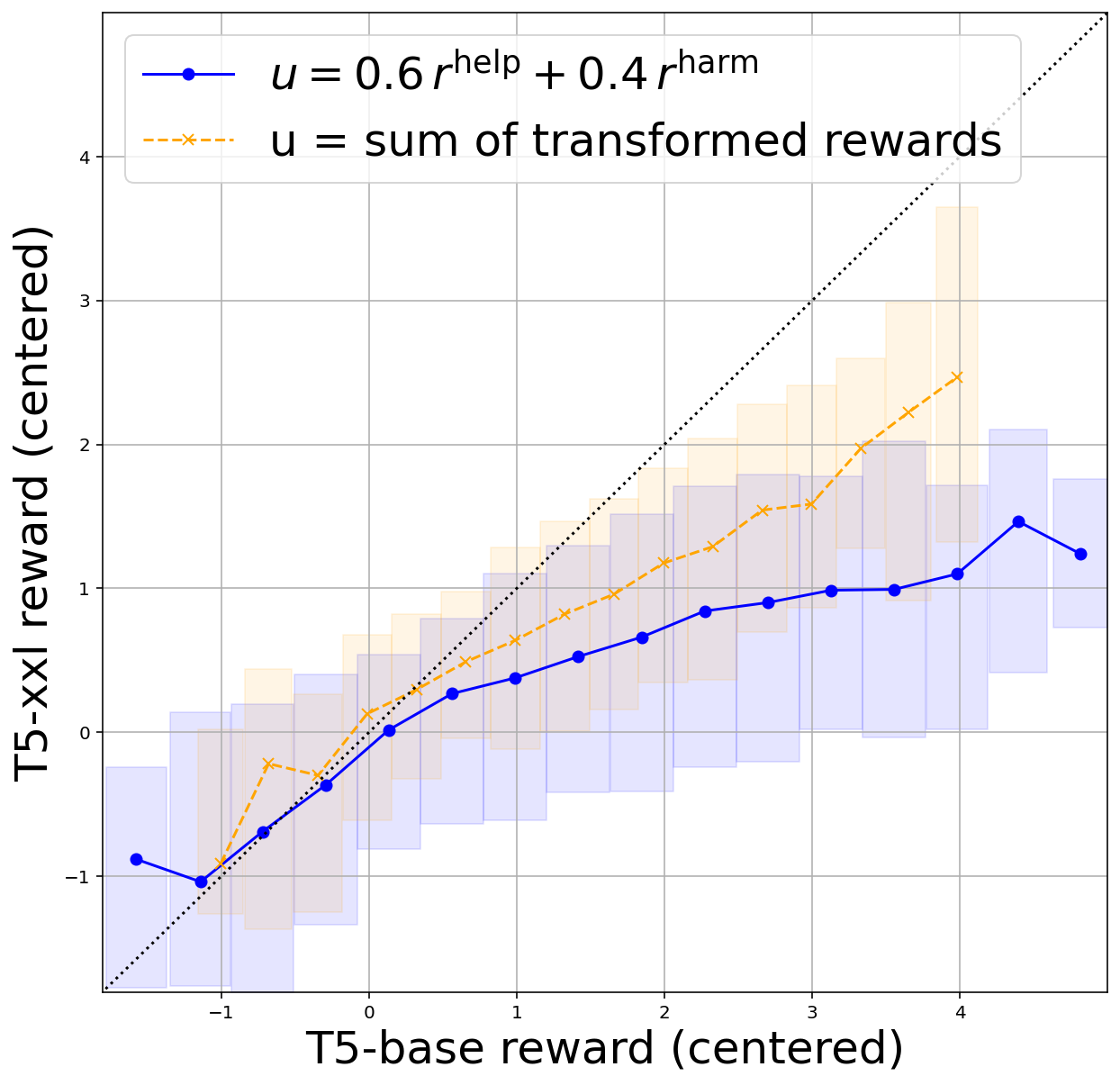}
        \caption{Helpful (aggregated reward): $\text{KL} \approx 14$}    
        \label{fig:help_base_vs_xxl_agg}
    \end{subfigure}
    \begin{subfigure}{0.48\textwidth}
        \centering
        \includegraphics[width=\linewidth]{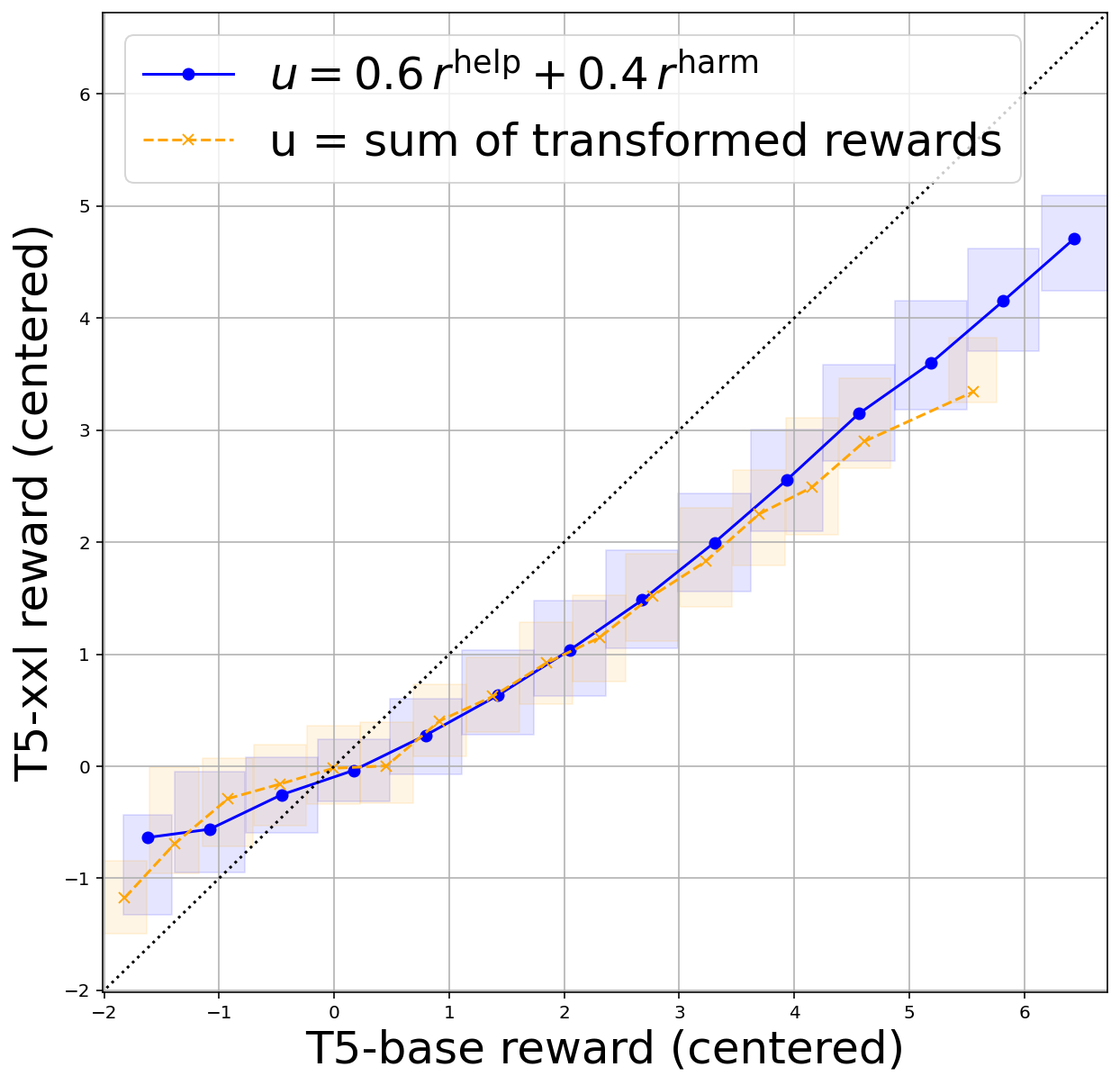}
        \caption{Harmless (aggregated reward): $\text{KL} \approx 14$}   
        \label{fig:harm_base_vs_xxl_agg}
    \end{subfigure}
    \caption{Reward transformation mitigates reward overoptimization, particularly for responses with larger reward values, and in reward aggregation where reward hacking is more severe. We compare reward overoptimization patterns in policies aligned with raw reward and transformed reward, in single reward (\cref{fig:help_base_vs_xxl_single}, \cref{fig:harm_base_vs_xxl_single}) and reward aggregation (\cref{fig:help_base_vs_xxl_agg}, \cref{fig:harm_base_vs_xxl_agg}) settings. The choice of aligned policies and score centering are the same as in \cref{fig:reward_distn}. For each plot, we sort the centered scores from T5-base reward model into $20$ equal-width bins. For each bin with more than $10$ data points: in the x-axis we show the interval medium; in the y-axis, we visualize the 25\%, 50\% and 75\% quantile of the corresponding centered scores from T5-xxl.}
    \label{fig:base_vs_xxl}
\end{figure*}

In \cref{fig:reward_distn} we choose a pair of aligned policies matched on KL values to show that reward transformation leads to more uniform improvement than using the raw reward.
In \cref{fig:base_vs_xxl} we visually show reward overoptimization (in the same aligned policies), and how reward transformation alleviates it. The mitigation is most obvious for responses with larger reward values. It's also important to note that reward aggregation exacerbates reward hacking with weighed sum of the raw reward: the policy model is incentivized to generate responses with very high harmlessness score (despite having smaller weight for harmlessness); to retain good scores in helpfulness, the policy model hacks the helpfulness reward model even more (compare \cref{fig:help_base_vs_xxl_single} against \cref{fig:help_base_vs_xxl_agg}). Using sum of transformed reward alleviates this by two mechanisms.

\begin{figure*}
    \centering
    \begin{subfigure}{.5\textwidth}
      \centering
      \includegraphics[width=\linewidth]{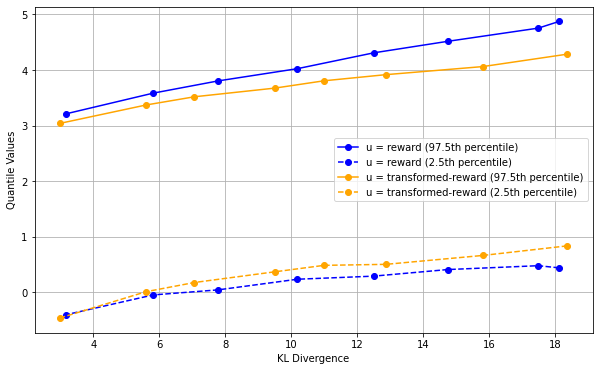}
      \caption{Helpfulness}
    \end{subfigure}%
    \centering
    \begin{subfigure}{.5\textwidth}
      \centering
      \includegraphics[width=\linewidth]{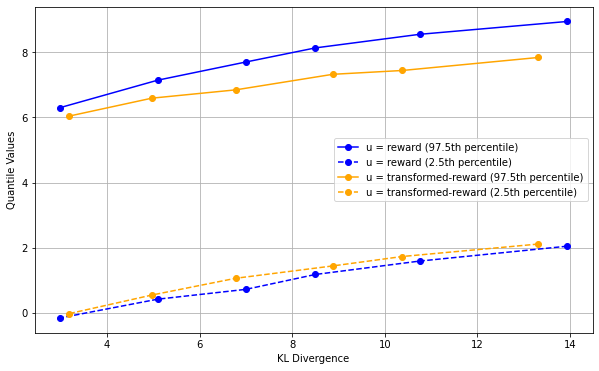}
      \caption{Harmlessness}
    \end{subfigure}%
    \caption{Reward Transformation leads to more uniform reward improvements than baseline. The reward values are centered by median SFT rewards. This complements the results in \cref{fig:reward_distn} for comparisons across KL values.}
    \label{fig:quantile_vs_kl}
\end{figure*}

In \cref{fig:quantile_vs_kl} we show reward transformation leads to more concentrated reward distribution than using raw reward, the same trend across different KL values. 

\begin{figure*}
  \centering
  \begin{subfigure}{\linewidth}
    \centering
    \begin{subfigure}{.25\textwidth}
      \centering
      \includegraphics[width=\linewidth]{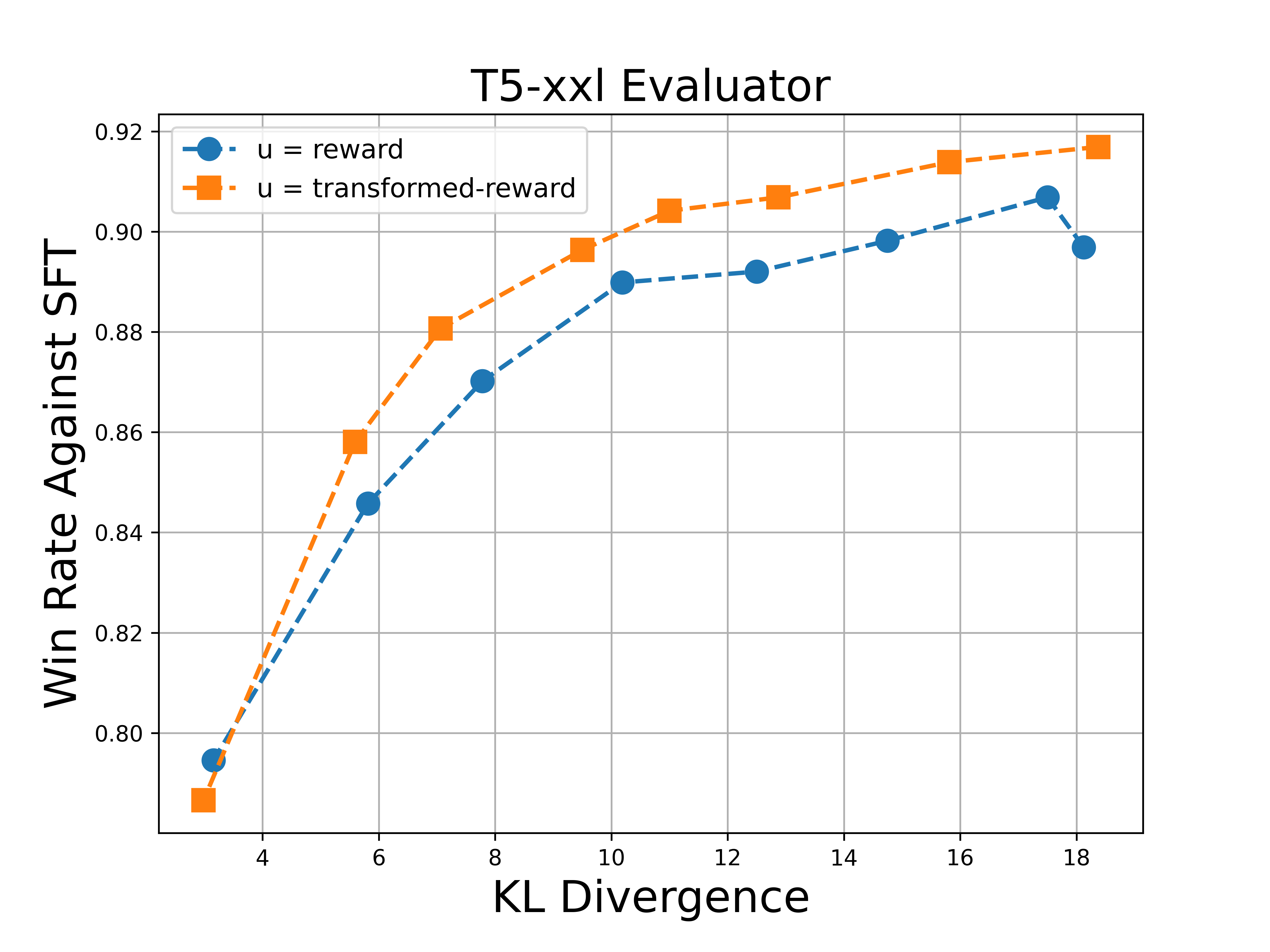}
      \caption{SFT Random Sample}
    \end{subfigure}%
    \begin{subfigure}{.25\textwidth}
      \centering
      \includegraphics[width=\linewidth]{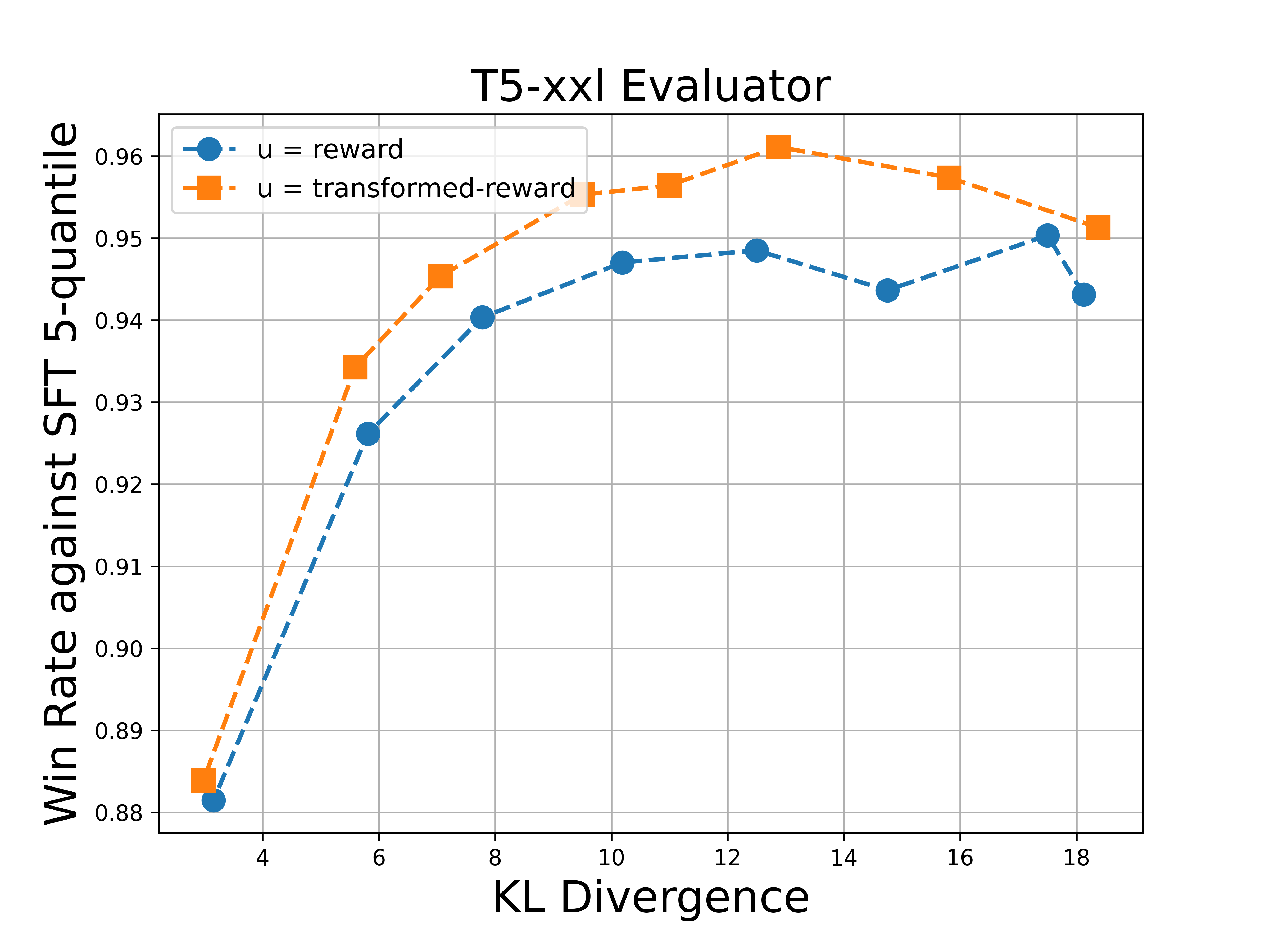}
      \caption{SFT 50\%-quantile}
    \end{subfigure}%
    \begin{subfigure}{.25\textwidth}
      \centering
      \includegraphics[width=\linewidth]{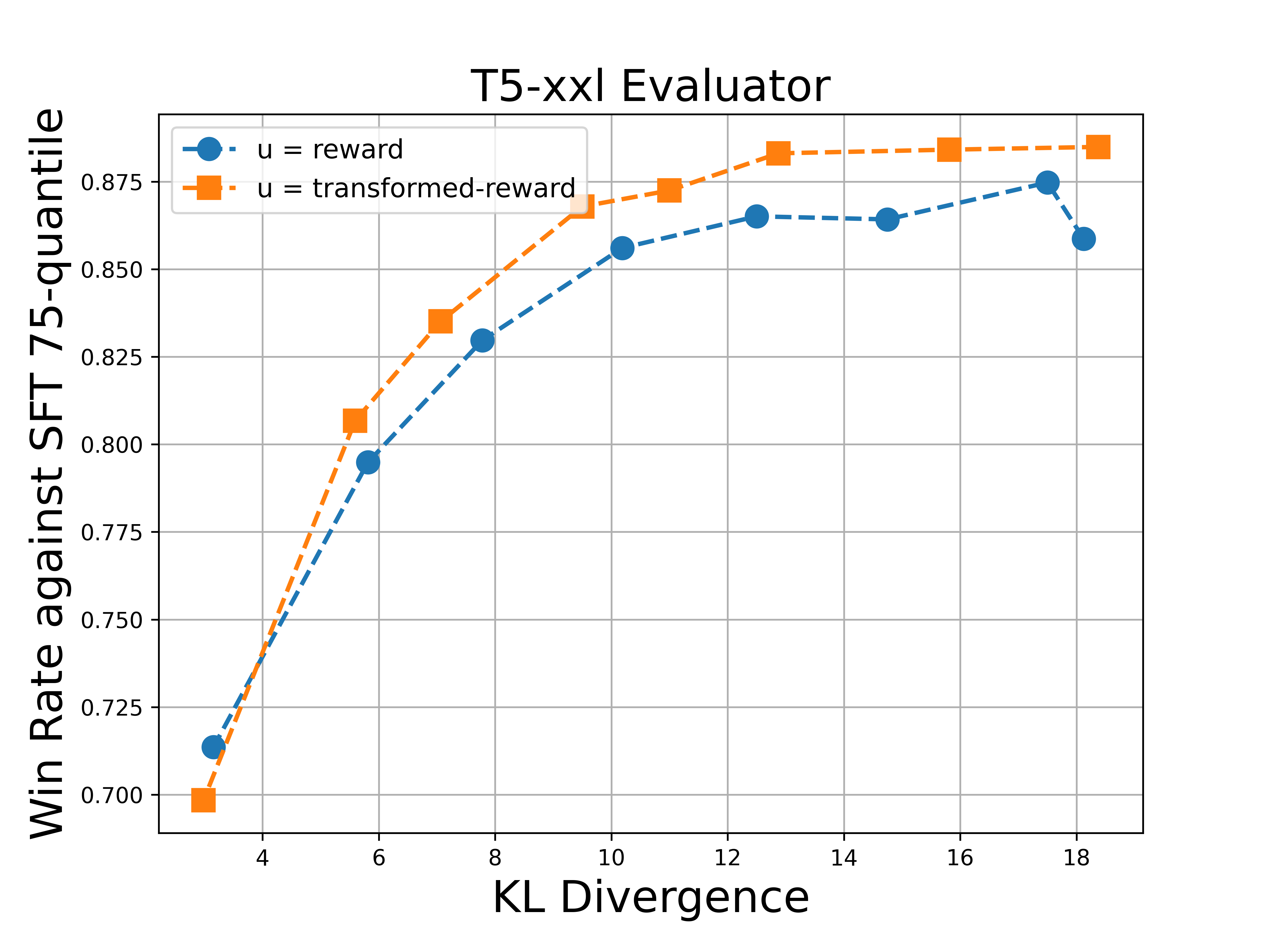}
      \caption{SFT 75\%-quantile}
    \end{subfigure}%
     \begin{subfigure}{.25\textwidth}
      \centering
      \includegraphics[width=\linewidth]{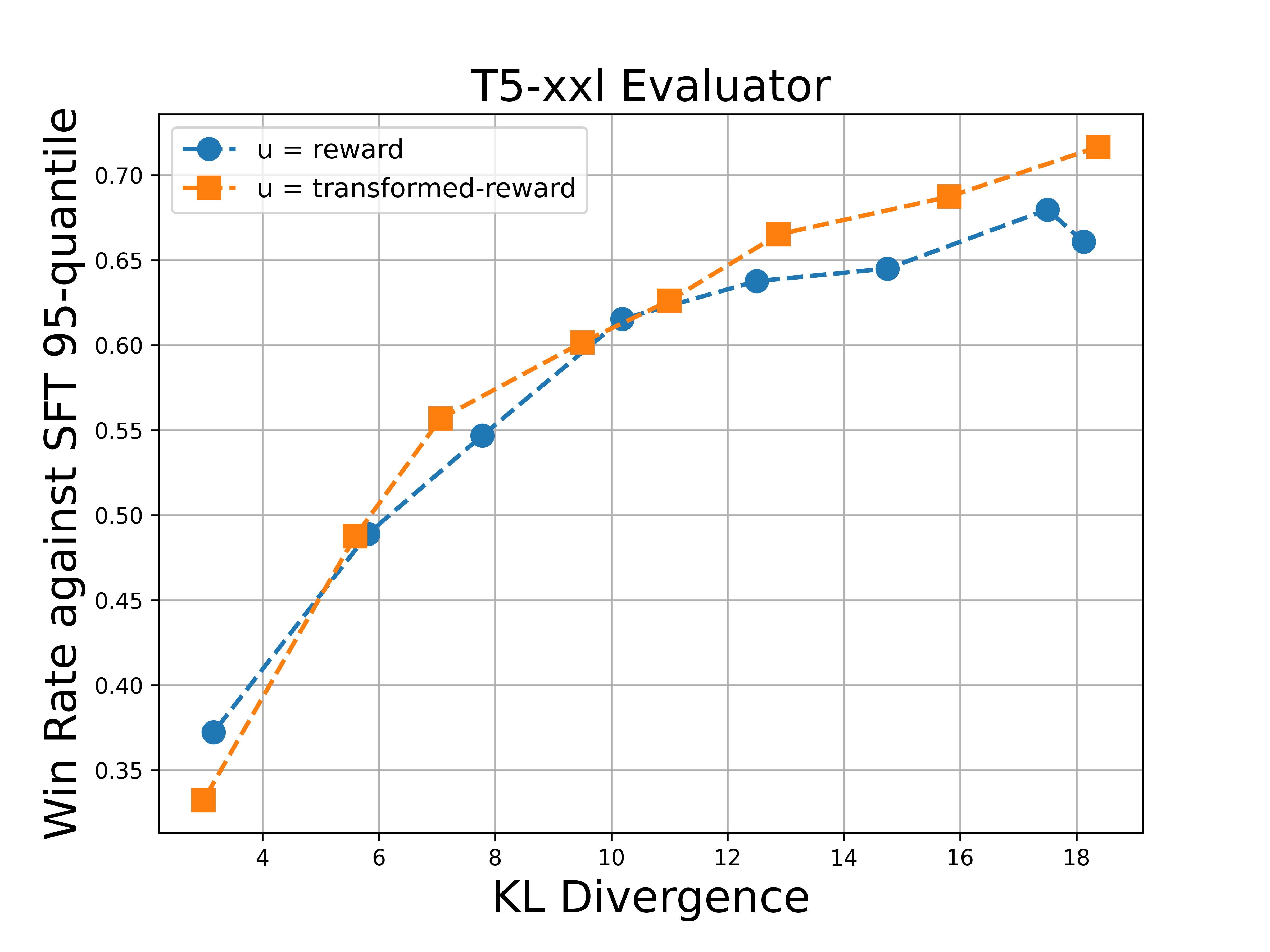}
      \caption{SFT 95\%-quantile}
    \end{subfigure}%
    \caption{Helpfulness}
  \end{subfigure}%

  \begin{subfigure}{\linewidth}
    \centering
    \begin{subfigure}{.25\textwidth}
      \centering
      \includegraphics[width=\linewidth]{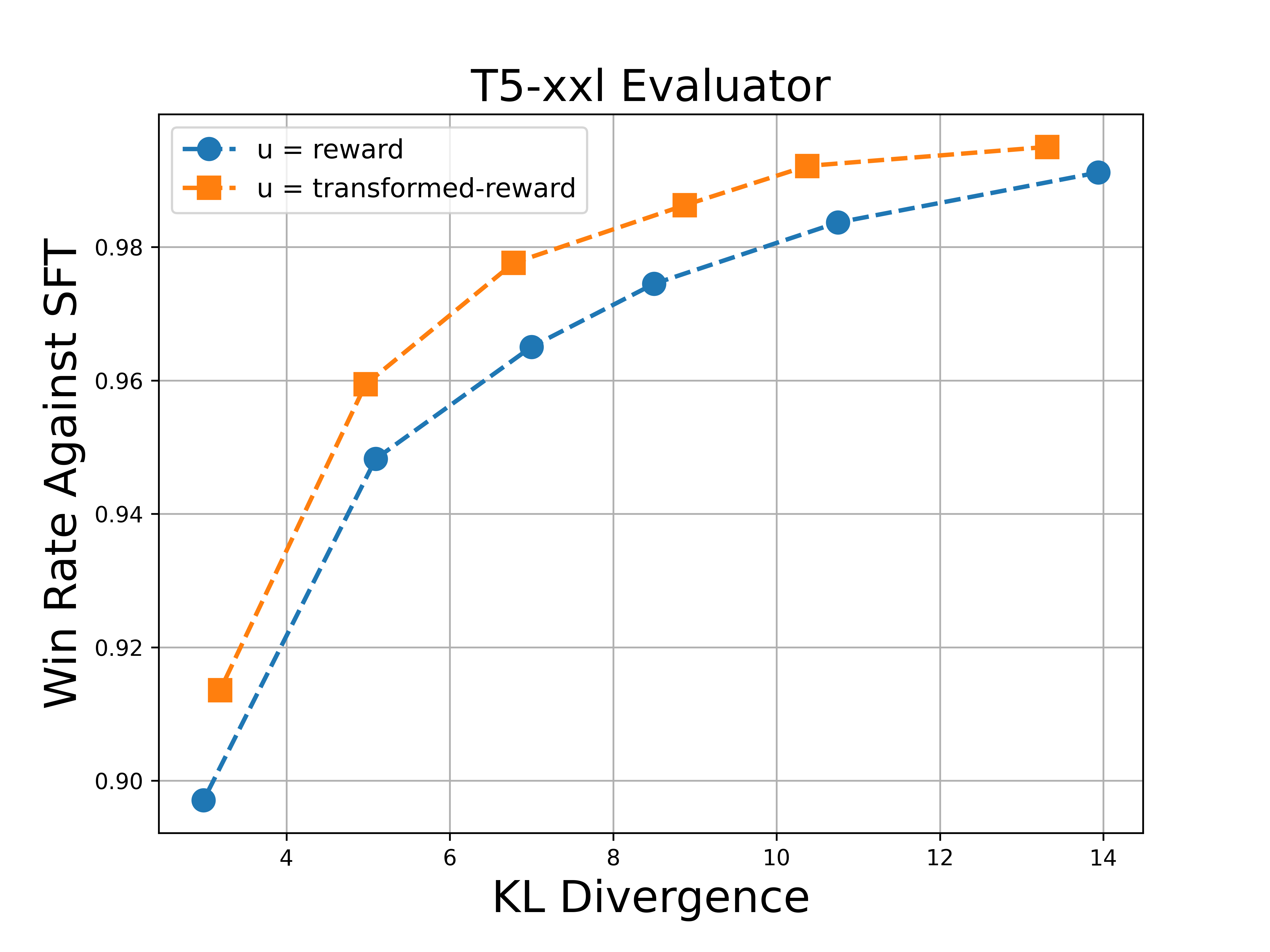}
      \caption{SFT Random Sample}
    \end{subfigure}%
    \begin{subfigure}{.25\textwidth}
      \centering
      \includegraphics[width=\linewidth]{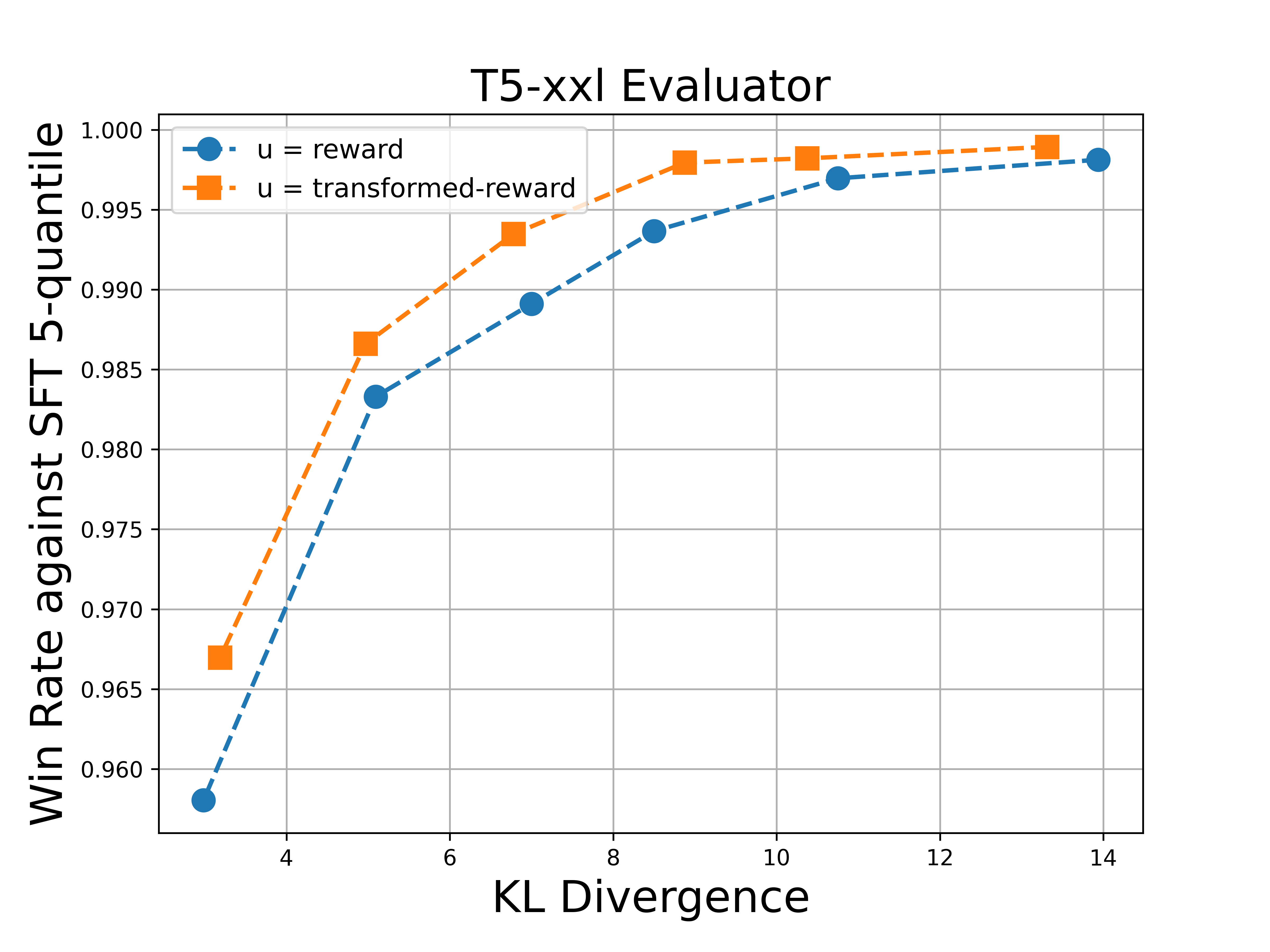}
      \caption{SFT 50\%-quantile}
    \end{subfigure}%
    \begin{subfigure}{.25\textwidth}
      \centering
      \includegraphics[width=\linewidth]{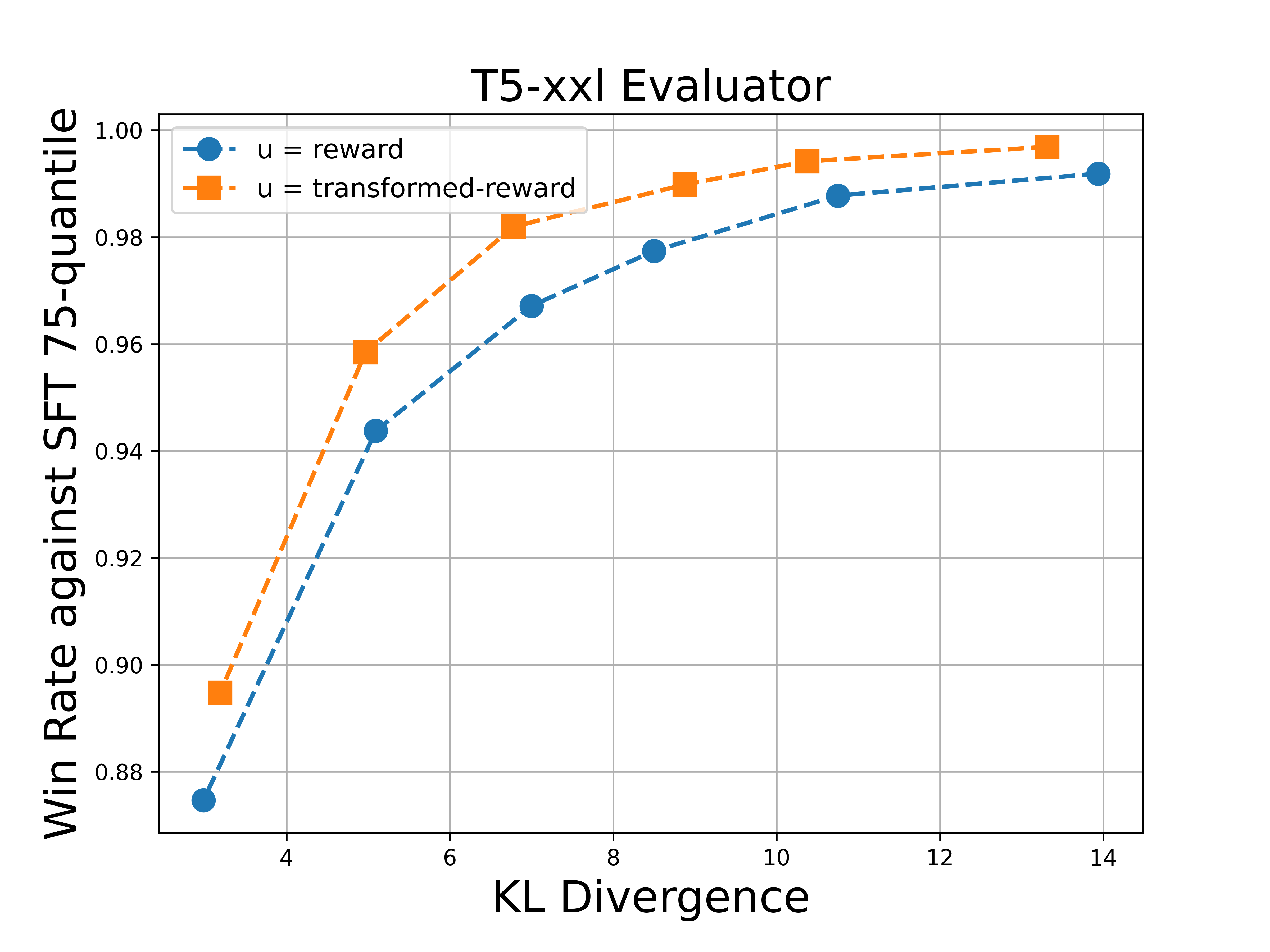}
      \caption{SFT 75\%-quantile}
    \end{subfigure}%
     \begin{subfigure}{.25\textwidth}
      \centering
      \includegraphics[width=\linewidth]{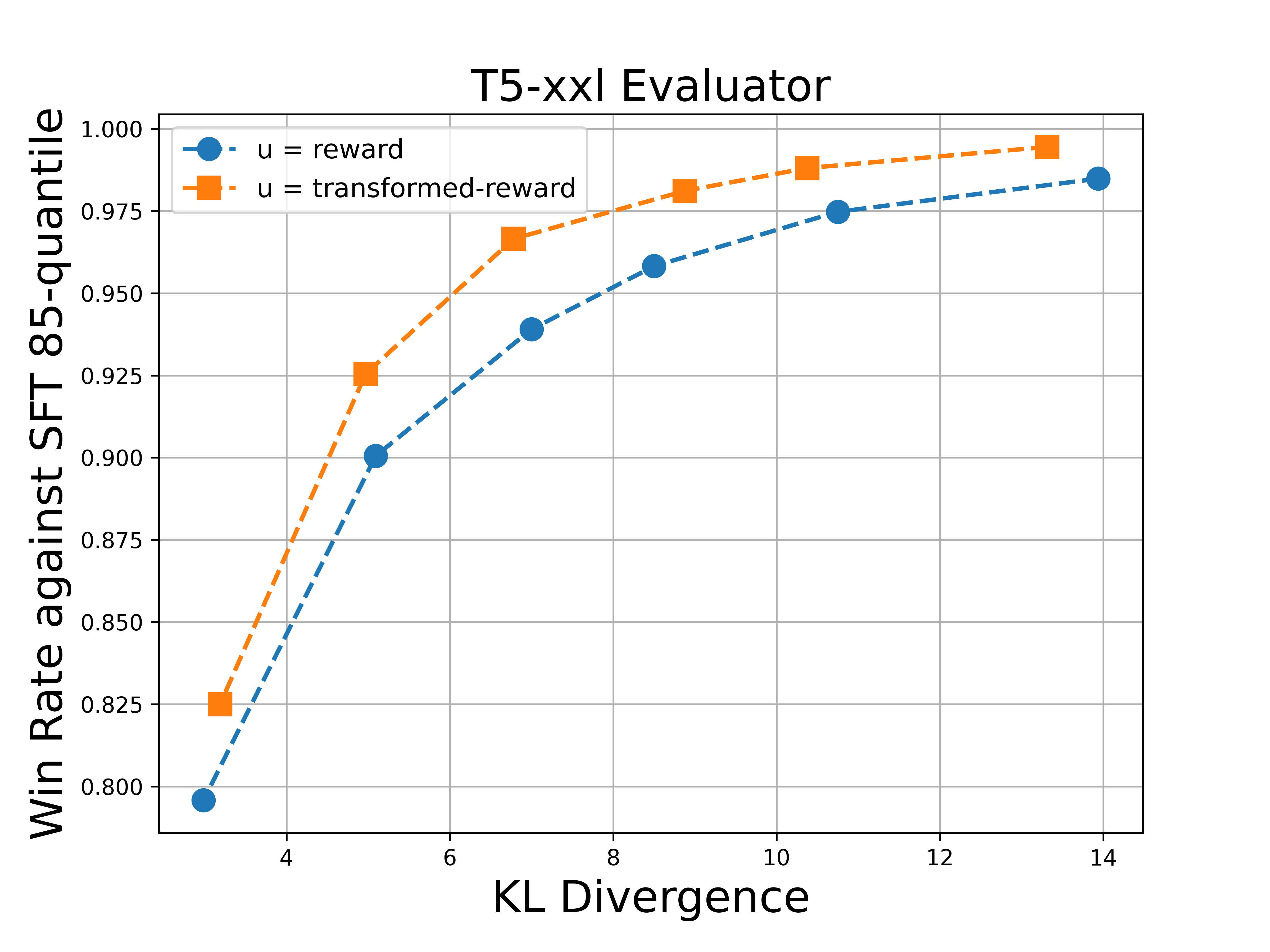}
      \caption{SFT 85\%-quantile}
    \end{subfigure}%
    \caption{Harmlessness}
  \end{subfigure}%
  \caption{Transformed reward obtains better KL and win-rate trade-offs. These are win-rates compared against SFT random sample, and $q$-th quantile of the rewards of SFT samples. See details in \cref{fig:win_rate_single}.}
  \label{fig:win_rate_single_extra}
\end{figure*}

In \cref{fig:win_rate_single} we report win-rate against random sample under PALM-2, and against SFT 85\%-quantile for helpfulness and 95\%-quantile for harmlessness. In \cref{fig:win_rate_single_extra}, we report the extra evaluations. 

\end{document}